\def\year{2020}\relax
\documentclass[letterpaper]{article} 
\usepackage{aaai20}  
\usepackage{times}  
\usepackage{helvet} 
\usepackage{courier}  
\usepackage[hyphens]{url}  
\usepackage{graphicx} 
\usepackage{graphicx}  
\usepackage{amssymb}
\usepackage{algorithmicx}
\usepackage[ruled]{algorithm}
\usepackage{algpseudocode}
\usepackage{enumerate}
\usepackage{comment}
\usepackage[tbtags]{amsmath}
\usepackage{xspace}
\usepackage{algorithmicx}
\usepackage[ruled]{algorithm}
\usepackage{algpseudocode}
\usepackage{url}
\usepackage{underscore}
\usepackage{multirow}
\usepackage{array}
\usepackage{color}

\floatname{algorithm}{}

\newcommand{\ie}{{i.e.}\xspace}
\newcommand{\eg}{{e.g.}\xspace}

\newcommand{\wrt}{{w.r.t.}\xspace}

\newcommand{\Tab}{Table\xspace}

\newcommand{\calL}{\mathcal{L}}

\title{Distance-IoU Loss: Faster and Better Learning for Bounding Box Regression }
\author{\textbf{Zhaohui Zheng}\textsuperscript{\rm 1},
	\textbf{Ping Wang}\textsuperscript{\rm 1},
	\textbf{Wei Liu}\textsuperscript{\rm 2},
	\textbf{Jinze Li}\textsuperscript{\rm 3},
	\textbf{Rongguang Ye}\textsuperscript{\rm 1},
	\textbf{Dongwei Ren*}\textsuperscript{\rm 2}\\ 
	\\
	\textsuperscript{\rm 1}School of Mathematics, Tianjin University, China\\ 
	\textsuperscript{\rm 2}College of Intelligence and Computing, Tianjin University, China\\
	\textsuperscript{\rm 3}School of Information Technology and Cyber Security, People's Public Security University of China\\
	\\
	*Corresponding author: rendongweihit@gmail.com 
}

\begin{document}
	
	\maketitle

	\begin{abstract}
		Bounding box regression is the crucial step in object detection.
		In existing methods, while $\ell_n$-norm loss is widely adopted for bounding box regression, it is not tailored to the evaluation metric, \ie, Intersection over Union (IoU).
		Recently, IoU loss and generalized IoU (GIoU) loss have been proposed to benefit the IoU metric, but still suffer from the problems of slow convergence and inaccurate regression.
		In this paper, we propose a Distance-IoU (DIoU) loss by incorporating the normalized distance between the predicted box and the target box, which converges much faster in training than IoU and GIoU losses.
		Furthermore, this paper summarizes three geometric factors in bounding box regression, \ie, overlap area, central point distance and aspect ratio, based on which a Complete IoU (CIoU) loss is proposed, thereby leading to faster convergence and better performance.
		By incorporating DIoU and CIoU losses into state-of-the-art object detection algorithms, \eg, YOLO v3, SSD and Faster R-CNN, we achieve notable performance gains in terms of not only IoU metric but also GIoU metric.
		Moreover, DIoU can be easily adopted into non-maximum suppression (NMS) to act as the criterion, further boosting performance improvement.
		%
        The source code and trained models are available at \url{https://github.com/Zzh-tju/DIoU}.
	\end{abstract}
	
	\noindent

	Object detection is one of the key issues in computer vision tasks, and has received considerable research attention for decades \cite{yolov1,yolov3,fasterrcnn,maskrcnn,MetaAnchor,wang2019data,Wang2018CVPR}.
	Generally, existing object detection methods can be categorized as: one-stage detection, such as YOLO series \cite{yolov1,yolov2,yolov3} and SSD \cite{SSD,DSSD}, two-stage detection, such as R-CNN series \cite{rcnn,fastrcnn,fasterrcnn,maskrcnn}, and even multi-stage detection, such as Cascade R-CNN \cite{cascadercnn}.
	Despite of these different detection frameworks, bounding box regression is the crucial step to predict a rectangular box to locate the target object.

	\begin{figure}[!t]
		\setlength{\abovecaptionskip}{0.cm}
		\setlength{\belowcaptionskip}{-0.cm}
		\centering
		\includegraphics[width=.45\textwidth]{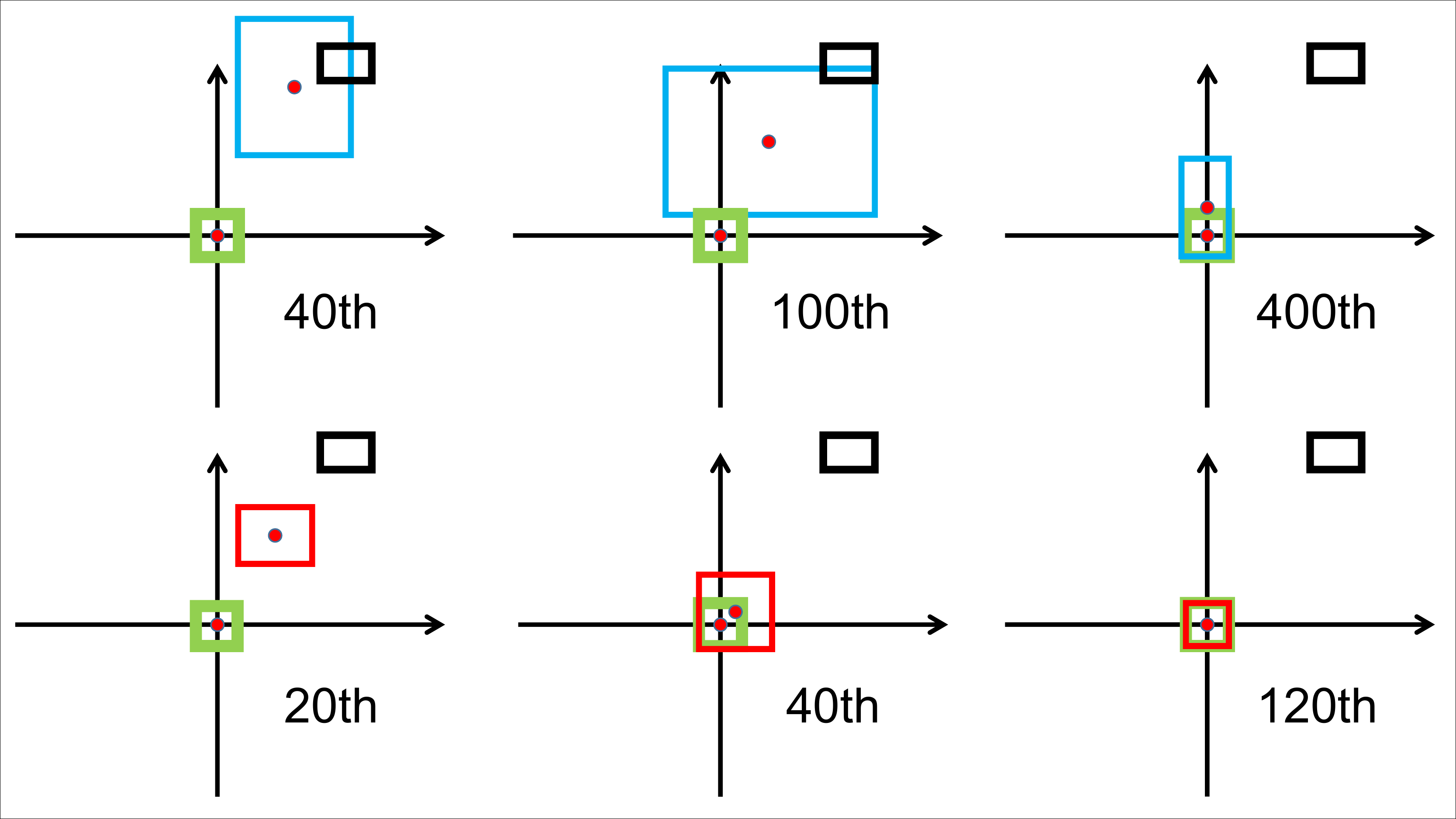}
		\caption{\footnotesize Bounding box regression steps by GIoU loss (first row) and DIoU loss (second row).
			{\color{green}{Green}} and \textbf{black} denote {\color{green}{target}} box and \textbf{anchor} box, respectively.
			{\color{blue}{Blue}} and {\color{red}{red}} denote predicted boxes for {\color{blue}{GIoU}} loss and {\color{red}{DIoU}} loss, respectively.
			GIoU loss generally increases the size of predicted box to overlap with target box, while DIoU loss directly minimizes normalized distance of central points. }
		\label{fig:regression steps}
	\end{figure}
	
	In terms of evaluation metric for bounding box regression, Intersection over Union (IoU) is the most popular metric,
	\begin{equation}\label{eq:iou}
	IoU = \frac{|B\cap B^{gt}|}{|B\cup B^{gt}|},
	\end{equation}
	where $B^{gt}=(x^{gt},y^{gt},w^{gt},h^{gt})$ is the ground-truth, and $B=(x,y,w,h)$ is the predicted box.
	Conventionally, $\ell_n$-norm (\eg, $n=1 \text{ or } 2$) loss is adopted on the coordinates of $B$ and $B^{gt}$ to measure the distance between bounding boxes \cite{yolov1,fastrcnn,fasterrcnn,maskrcnn,RDAD}.
	However, as suggested in \cite{unitbox,giou}, $\ell_n$-norm loss is not a suitable choice to obtain the optimal IoU metric.
	In \cite{giou}, IoU loss is suggested to be adopted for improving the IoU metric,
	\begin{equation}\label{eq:iou loss}
	\mathcal{L}_{IoU}=1-\frac{|B\cap B^{gt}|}{|B\cup B^{gt}|}.
	\end{equation}
	However, IoU loss only works when the bounding boxes have overlap, and would not provide any moving gradient for non-overlapping cases.
	And then generalized IoU loss (GIoU) \cite{giou} is proposed by adding a penalty term,
	\begin{equation}\label{eq:giou loss}
	\mathcal{L}_{GIoU}=1- IoU + \frac{|C-B\cup B^{gt}|}{|C|},
	\end{equation}
	where $C$ is the smallest box covering $B$ and $B^{gt}$.
	Due to the introduction of penalty term, the predicted box will move towards the target box in non-overlapping cases.

\begin{figure}[!tb]
		\footnotesize
		\setlength{\abovecaptionskip}{0.cm}
		\setlength{\belowcaptionskip}{-0.cm}
		\centering
		\begin{tabular}{cccccc}
			\includegraphics[width=0.1\textwidth]{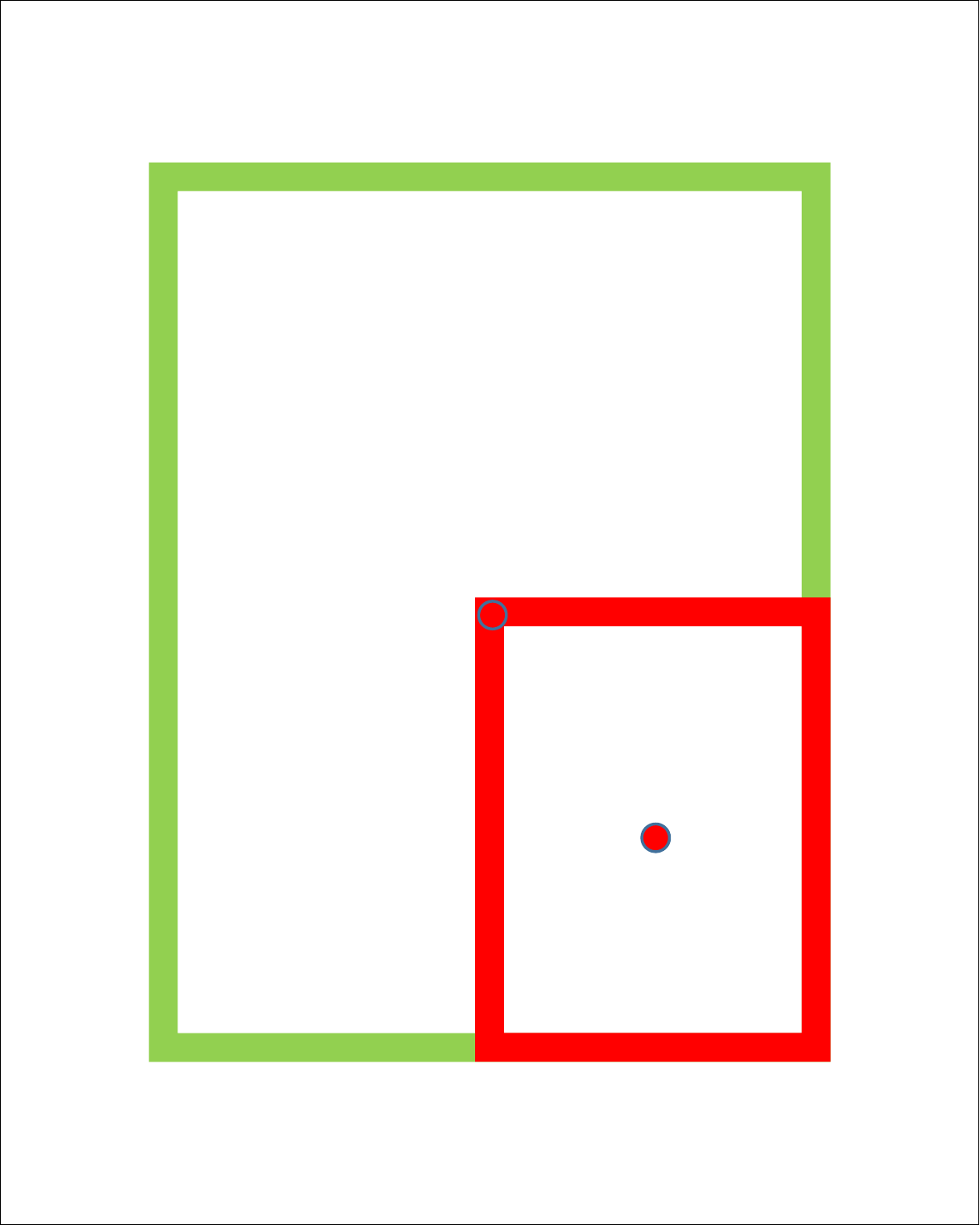}&
			\includegraphics[width=0.1\textwidth]{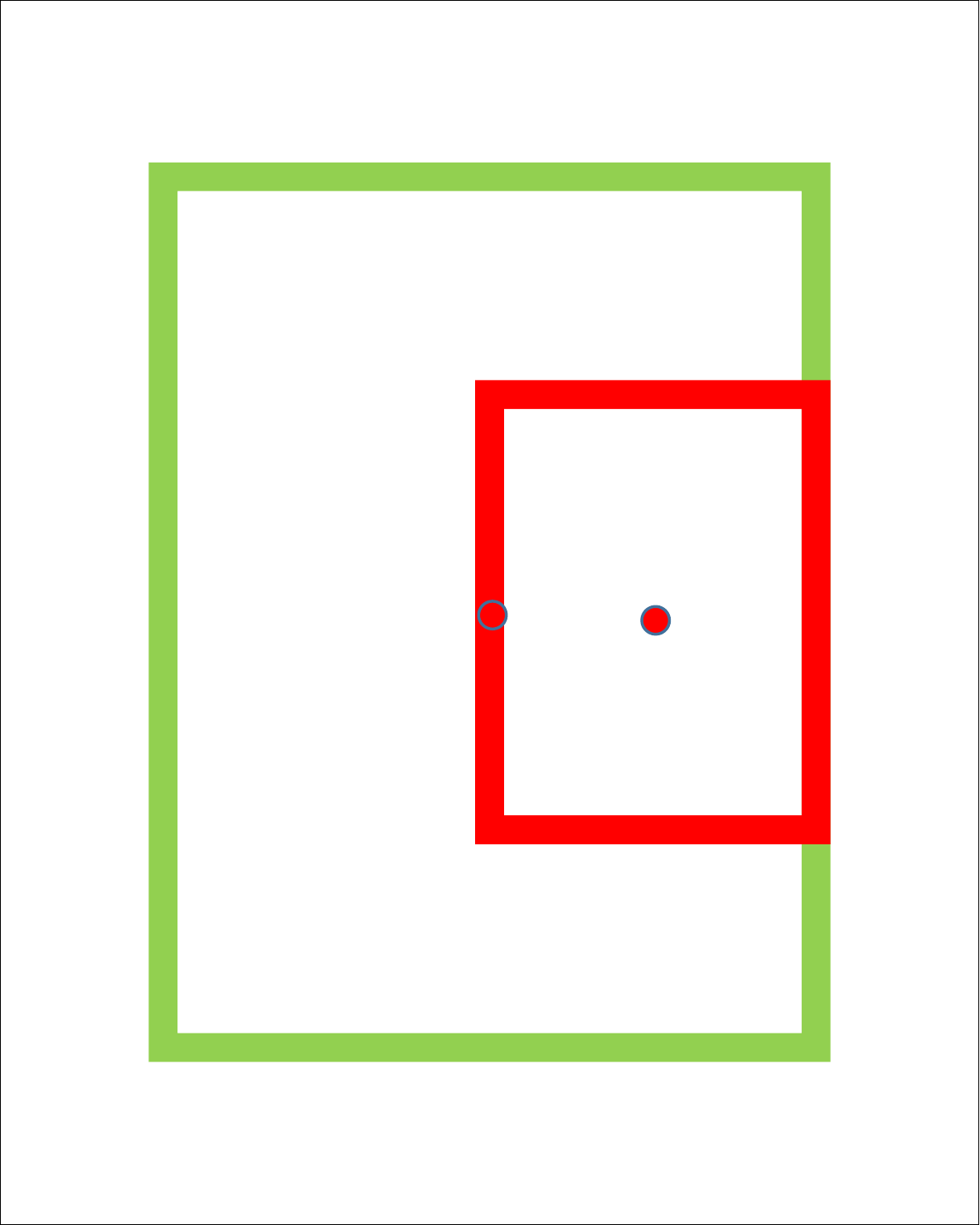}&
			\includegraphics[width=0.1\textwidth]{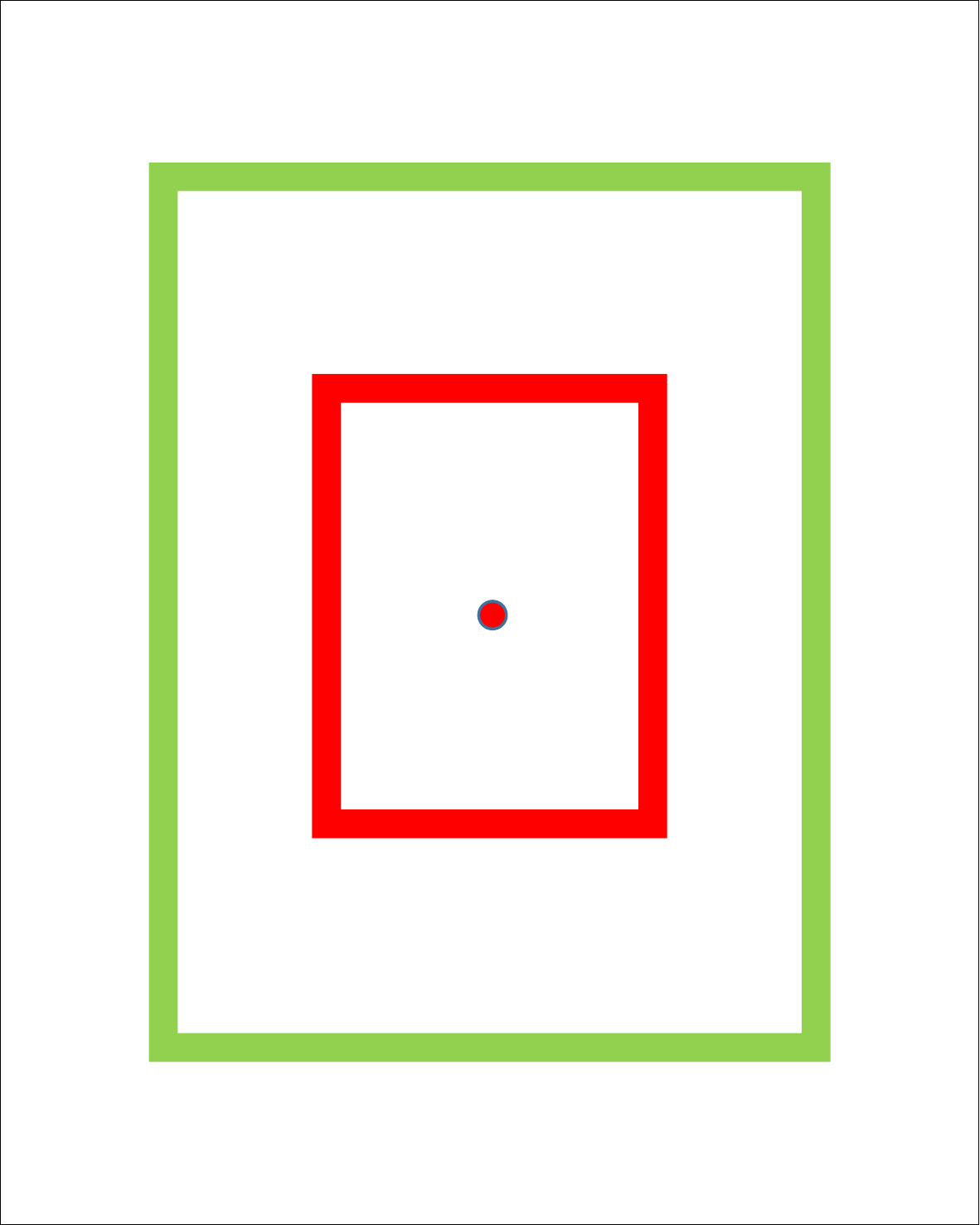}\\
			$\mathcal{L}_{IoU}=0.75$ & $\mathcal{L}_{IoU}=0.75$ & $\mathcal{L}_{IoU}=0.75$\\
			$\mathcal{L}_{GIoU}=0.75$ & $\mathcal{L}_{GIoU}=0.75$ & $\mathcal{L}_{GIoU}=0.75$\\
			$\mathcal{L}_{DIoU}=0.81$ & $\mathcal{L}_{DIoU}=0.77$ & $\mathcal{L}_{DIoU}=0.75$\\
		\end{tabular}
		\caption{\footnotesize GIoU loss degrades to IoU loss for these cases, while our DIoU loss is still distinguishable. {\color{green}{Green}} and {\color{red}{red}} denote {\color{green}{target}} box and {\color{red}{predicted}} box respectively.}
		\label{fig:loss example}
	\end{figure}

	Although GIoU can relieve the gradient vanishing problem for non-overlapping cases, it still has several limitations.
	By a simulation experiment (see \emph{Sec. \textbf{Analysis to IoU and GIoU Losses}} for details), we can evaluate the performance of GIoU loss for various bounding box positions.
	As shown in Fig. \ref{fig:regression steps}, one can see that GIoU loss intends to increase the size of predicted box at first, making it have overlap with target box, and then the IoU term in Eqn. \eqref{eq:giou loss} will work to maximize the overlap area of bounding box.
	And from Fig. \ref{fig:loss example}, GIoU loss will totally degrade to IoU loss for enclosing bounding boxes.
	Due to heavily relying on the IoU term, GIoU empirically needs more iterations to converge, especially for horizontal and vertical bounding boxes (see Fig. \ref{fig:finalerror}).
	Usually GIoU loss cannot well converge in the state-of-the-art detection algorithms, yielding inaccurate detection.

	%

	In this paper, we propose a Distance-IoU (DIoU) loss for bounding box regression.
	In particular, we simply add a penalty term on IoU loss to directly minimize the normalized distance between central points of two bounding boxes, leading to much faster convergence than GIoU loss.
	From Fig. \ref{fig:regression steps}, DIoU loss can be deployed to directly minimize the distance between two bounding boxes.
	And with only 120 iterations, the predicted box matches with the target box perfectly, while GIoU does not converge even after 400 iterations.
	Furthermore, we suggest that a good loss for bounding box regression should consider three important geometric measures, \ie, overlap area, central point distance and aspect ratio, which have been ignored for a long time.
	By combining these geometric measures, we further propose a Complete IoU (CIoU) loss for bounding box regression, leading to faster convergence and better performance than IoU and GIoU losses.
	The proposed losses can be easily incorporated into the state-of-the-art object detection algorithms.
	%
	Moreover, DIoU can be employed as a criterion in non-maximum suppression (NMS), by which not only the overlap area but also the distance between central points of two bounding boxes are considered when suppressing redundant boxes, making it more robust for the cases with occlusions.

	To evaluate our proposed methods, DIoU loss and CIoU loss are incorporated into several state-of-the-art detection algorithms including YOLO v3 \cite{yolov3}, SSD \cite{SSD} and Faster R-CNN \cite{fasterrcnn}, and are evaluated on two popular benchmark datasets PASCAL VOC 2007 \cite{voc} and MS COCO 2017 \cite{coco}.

	The contribution of work is summarized as follows:
	\begin{enumerate}
		\item A Distance-IoU loss, \ie, DIoU loss, is proposed for bounding box regression, which has faster convergence than IoU and GIoU losses.
		
		\item A Complete IoU loss, \ie, CIoU loss, is further proposed by considering three geometric measures, \ie, overlap area, central point distance and aspect ratio, which better describes the regression of rectangular boxes.
		
		\item DIoU is deployed in NMS, and is more robust than original NMS for suppressing redundant boxes.
		
		\item The proposed methods can be easily incorporated into the state-of-the-art detection algorithms, achieving notable performance gains.
		
	\end{enumerate}
	
	\section{Related Work}
	\label{related work}
	
	In this section, we briefly survey relevant works including object detection methods, loss function for bounding box regression and non-maximum suppression.
	
	\subsection{Object Detection}
	In \cite{TLL}, the central axis line is applied in pedestrian detection.
	CornerNet \cite{cornernet} suggested predicting a pair of corners to replace a rectangular box for locating object.
	In RepPoints \cite{RepPoints}, a rectangular box is formed by predicting several points.
	Recently, FSAF \cite{FSAF} proposed anchor-free branch to tackle the issues of non-optimality in online feature selection.
	There are also several loss functions for object detection, \eg, focal loss \cite{focalloss}, class-balanced loss \cite{classbalancedloss}, balanced loss for classification and bounding box regression \cite{librarcnn}, and gradient flow balancing loss \cite{GHM}. 
	Nevertheless, the regression of rectangular boxes is still the most popular manner in the state-of-the-art object detection algorithms \cite{yolov3,maskrcnn,DSSD,SSD,FCOS}.
	\subsection{Loss Function for Bounding Box Regression}
	
	The $\ell_n$-norm loss functions are usually adopted in bounding box regression, but are sensitive to variant scales.
	In YOLO v1 \cite{yolov1}, square roots for $w$ and $h$ are adopted to mitigate this effect, while YOLO v3 \cite{yolov3} uses $2-wh$.
	%
	%
	%
	IoU loss is also used since Unitbox \cite{unitbox}, which is invariant to the scale.
	GIoU \cite{giou} loss is proposed to tackle the issues of gradient vanishing for non-overlapping cases, but is still facing the problems of slow convergence and inaccurate regression.
	In comparison, we propose DIoU and CIoU losses with faster convergence and better regression accuracy.
	
	\subsection{Non-Maximum Suppression}
	NMS is the last step in most object detection algorithms, in which redundant detection boxes are removed as long as its overlap with the highest score box exceeds a threshold.
	Soft-NMS \cite{softnms} penalizes the detection score of neighbors by a continuous function \wrt IoU, yielding softer and more robust suppression than original NMS.
	IoU-Net \cite{iounet} introduces a new network branch to predict the localization confidence to guide NMS.
	Recently, adaptive NMS \cite{adaptivenms} and Softer-NMS \cite{softernms} are proposed to respectively study proper threshold and weighted average strategies.
	%
	%
	In this work, DIoU is simply deployed as the criterion in original NMS, in which the overlap area and the distance between two central points of bounding boxes are simultaneously considered when suppressing redundant boxes.

	\section{Analysis to IoU and GIoU Losses} \label{sec:simulation}
	
	To begin with, we analyze the limitations of original IoU loss and GIoU loss.
	However, it is very difficult to analyze the procedure of bounding box regression simply from the detection results, where the regression cases in uncontrolled benchmarks are often not comprehensive, \eg, different distances, different scales and different aspect ratios.
	Instead, we suggest conducting simulation experiments, where the regression cases should be comprehensively considered, and then the issues of a given loss function can be easily analyzed.
	\begin{algorithm}[!tb]	
		
		\footnotesize
		\caption{\textbf{Simulation Experiment}}
		\label{algo:RSValue}
		\begin{algorithmic}[1]
			\small{	
				\Require{Loss $\mathcal{L}$ is a continuous bounded function defined on $\mathbb{R}^4_{+}$.
					\newline $\mathbb{M}=\{\{B_{n,s}\}_{s=1}^{S}\}_{n=1}^{N}$ is the set of anchor boxes at $N=5,000$ uniformly scattered points within the circular region with center $(10,10)$ and radius $3$, and $S=7\times 7$ covers $7$ scales and $7$ aspect ratios of anchor boxes.
					\newline $\mathbb{M}^{gt}=\{B_{i}^{gt}\}_{i=1}^{7}$ is the set of target boxes that are fixed at $(10,10)$ with area 1, and have $7$ aspect ratios.
				}
				\Ensure{Regression error $\mathbf{E}\in \mathbb{R}^{T\times N}$}}
			\State Initialize $\mathbf{E}=\mathbf{0}$ and maximum iteration $T$.
			\State{Do bounding box regression:}
			\For {$n=1$ to $N$}
			\For {$s=1$ to $S$}
			\For {$i=1$ to $7$}
			\For {$t=1$ to $T$}
			\State{$\eta = \begin{cases}0.1 &\text{if}\quad t<=0.8T \\0.01 &\text{if}\quad 0.8T<t<=0.9T \\0.001 &\text{if}\quad t>0.9T\end{cases}$}
			\State{\!\!\!\! $\nabla B_{n,s}^{t-1}$ is gradient of $\mathcal{L}(B_{n,s}^{t-1},B^{gt}_i)$ \wrt $B_{n,s}^{t-1}$}
			\State{$B_{n,s}^{t}=B_{n,s}^{t-1}+\eta(2-IoU_{n,s}^{t-1})\nabla B_{n,s}^{t-1}$}
			\State{$\mathbf{E}(t,n)=\mathbf{E}(t,n) +|B_{n,s}^{t}-B_i^{gt}|$}
			\EndFor
			\EndFor
			
			\EndFor
			\EndFor
			\State{\Return{$\mathbf{E}$}}
		\end{algorithmic}
	\end{algorithm}
	\subsection{Simulation Experiment}
\begin{figure}[!htb]
		\footnotesize
		\setlength{\abovecaptionskip}{0.cm}
		\setlength{\belowcaptionskip}{-0.cm}
		\centering
			\includegraphics[width=0.22\textwidth]{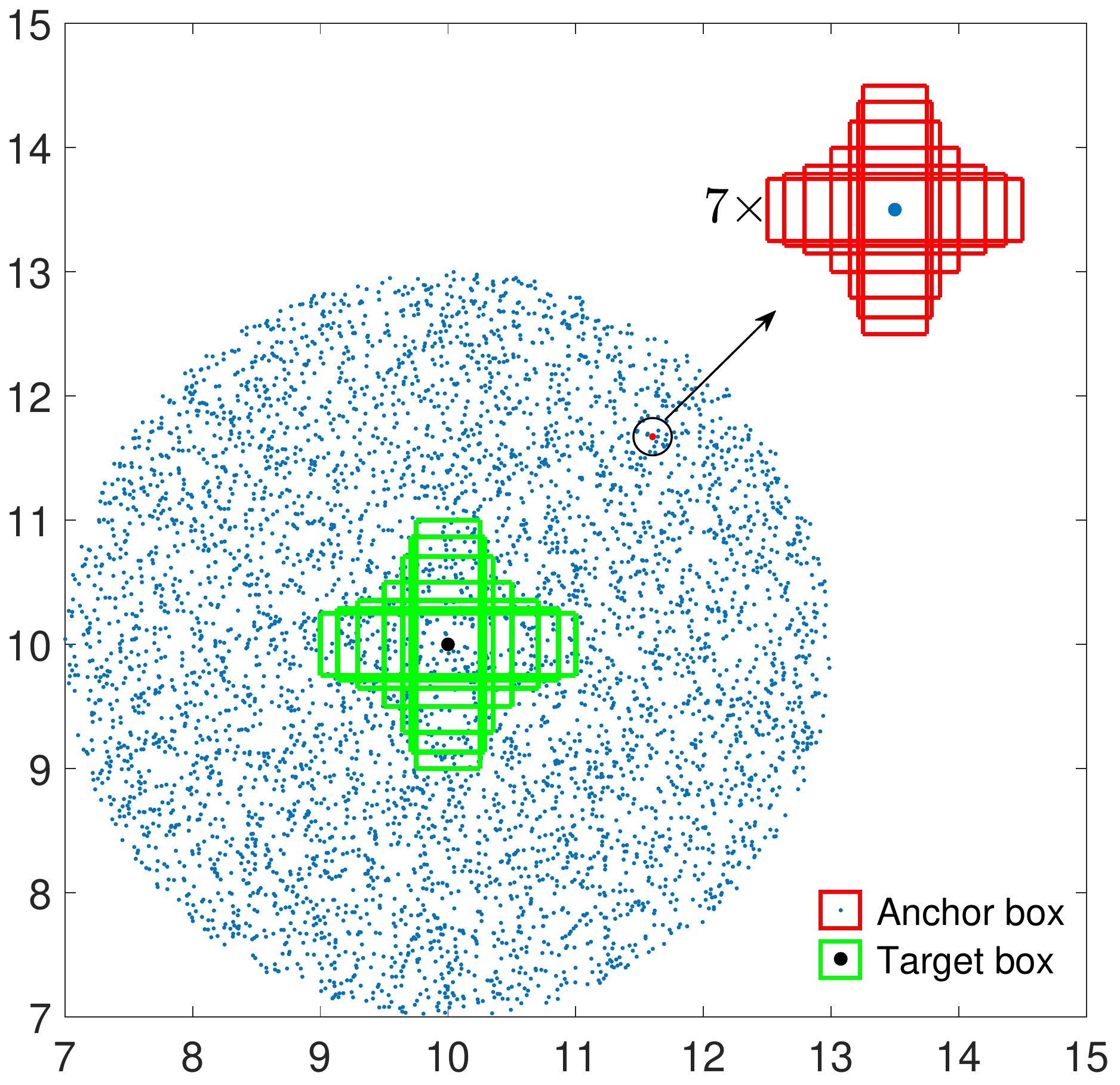}
			\includegraphics[width=0.24\textwidth]{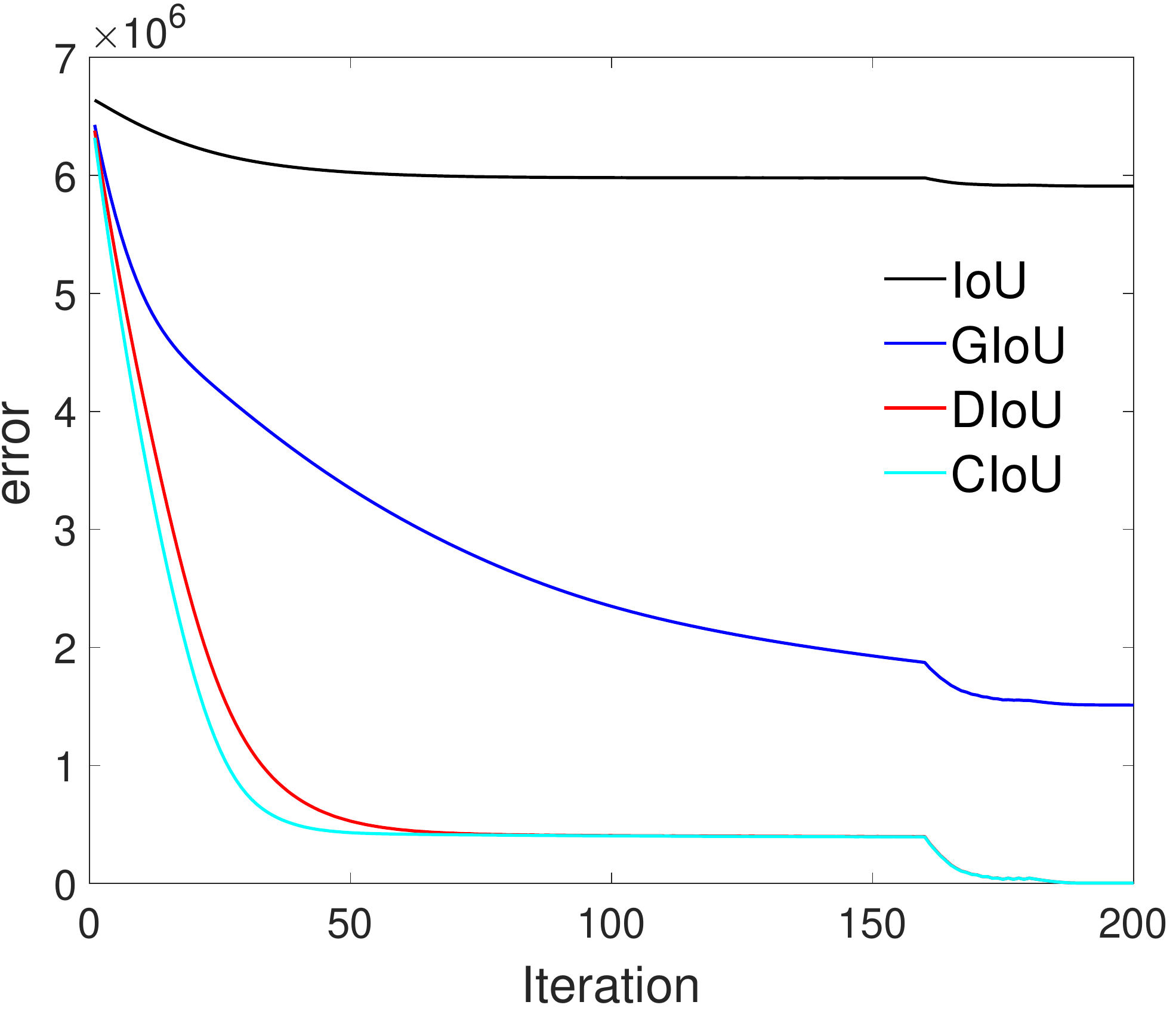}
		\caption{\footnotesize Simulation experiments: (a) 1,715,000 regression cases are adopted by considering different distances, scales and aspect ratios, (b) regression error sum (\ie, $\sum_{n} \mathbf{E}(t,n)$) curves of different loss functions at iteration $t$. }
		\label{fig:1715ksampling}
	\end{figure}

	In the simulation experiments, we try to cover most of the relationships between bounding boxes in terms of distance, scale and aspect ratio, as shown in Fig. \ref{fig:1715ksampling}(a).
	In particular, we choose 7 unit boxes (\ie, the area of each box is 1) with different aspect ratios (\ie, 1:4, 1:3, 1:2, 1:1, 2:1, 3:1 and 4:1) as target boxes.
	Without loss of generality, the central points of the 7 target boxes are fixed at $(10,10)$.
	The anchor boxes are uniformly scattered at 5,000 points.
	(\textbf{\emph{i}}) Distance: In the circular region centered at $(10,10)$ with radius 3, 5,000 points are uniformly chosen to place anchor boxes with 7 scales and 7 aspect ratios.
	In these cases, overlapping and non-overlapping boxes are included.
	(\textbf{\emph{ii}}) Scale: For each point, the areas of anchor boxes are set as $0.5$, $0.67$, $0.75$, $1$, $1.33$, $1.5$ and $2$.
	(\textbf{\emph{iii}}) Aspect ratio: For a given point and scale, 7 aspect ratios are adopted, \ie, following the same setting with target boxes (\ie, 1:4, 1:3, 1:2, 1:1, 2:1, 3:1 and 4:1).
	All the $5,000 \times 7 \times 7 $ anchor boxes should be fitted to each target box.
	To sum up, there are totally $1,715,000 = 7 \times 7 \times 7 \times 5,000$ regression cases.
	
	Then given a loss function $\mathcal{L}$, we can simulate the procedure of bounding box regression for each case using gradient descent algorithm.
	For predicted box $B_i$, the current prediction can be obtained by
	\begin{equation}
	\begin{aligned}
	B_{i}^{t}=B_{i}^{t-1}+\eta(2-IoU_{i}^{t-1})\nabla B_{i}^{t-1},
	\end{aligned}
	\end{equation}
	where $B_i^t$ is the predicted box at iteration $t$, $\nabla B_i^{t-1}$ denotes the gradient of loss $\mathcal{L}$ \wrt $B_i$ at iteration $t-1$, and $\eta$ is the step.
	It is worth noting that in our implementation, the gradient is multiplied by $2-IoU_i^{t-1}$ to accelerate the convergence.
	The performance of bounding box regression is evaluated using $\ell_1$-norm.
	%
	%
	For each loss function, the simulation experiment is terminated when reaching iteration $T=200$, and the error curves are shown in Fig. \ref{fig:1715ksampling}(b).

	\subsection{Limitations of IoU and GIoU Losses}
\begin{figure*}[!tb]
		\footnotesize
		\setlength{\abovecaptionskip}{0.cm}
		\setlength{\belowcaptionskip}{-0.cm}
		\centering
		\begin{tabular}{cccccccccc}
			\includegraphics[width=0.28\textwidth]{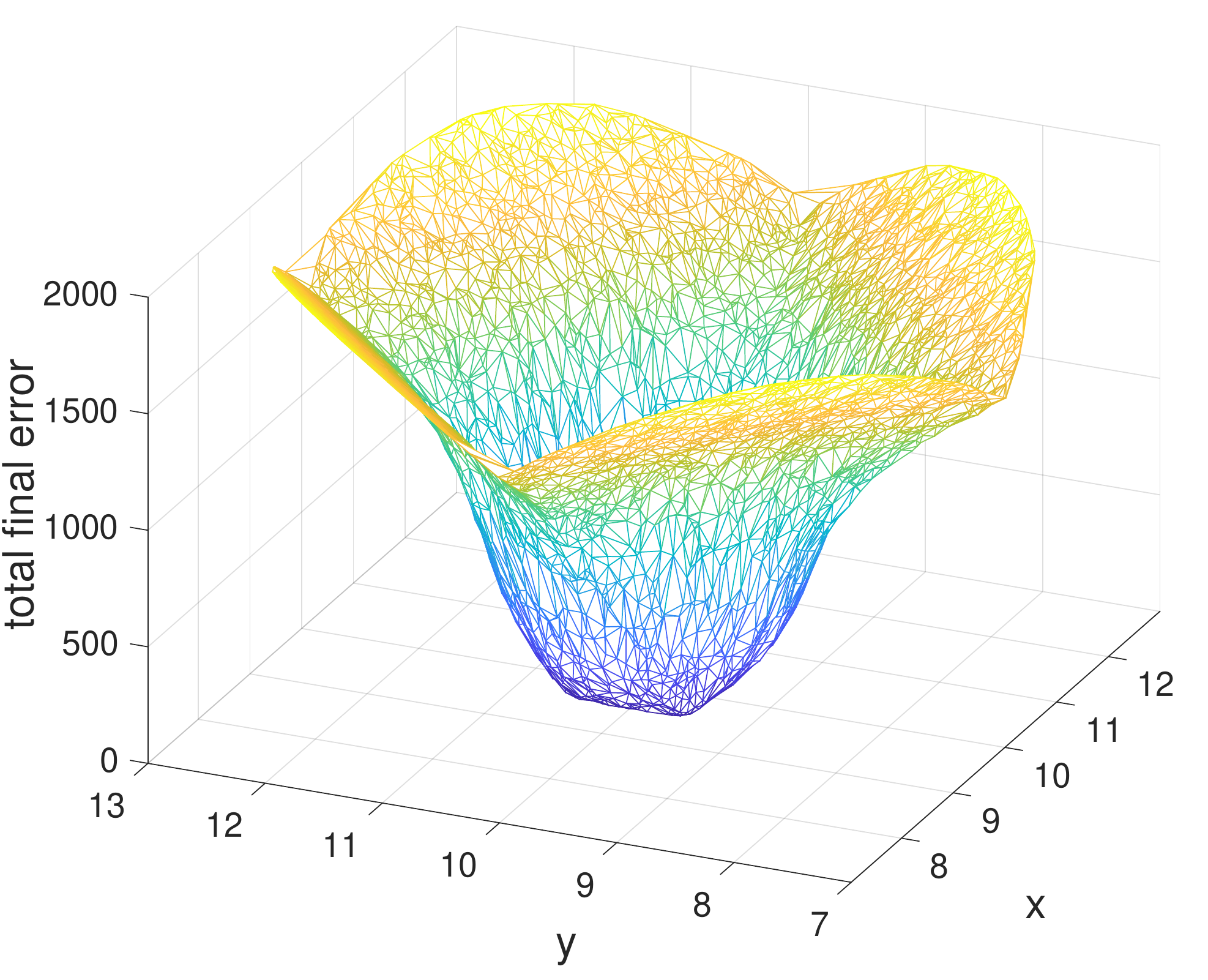}&
			\includegraphics[width=0.28\textwidth]{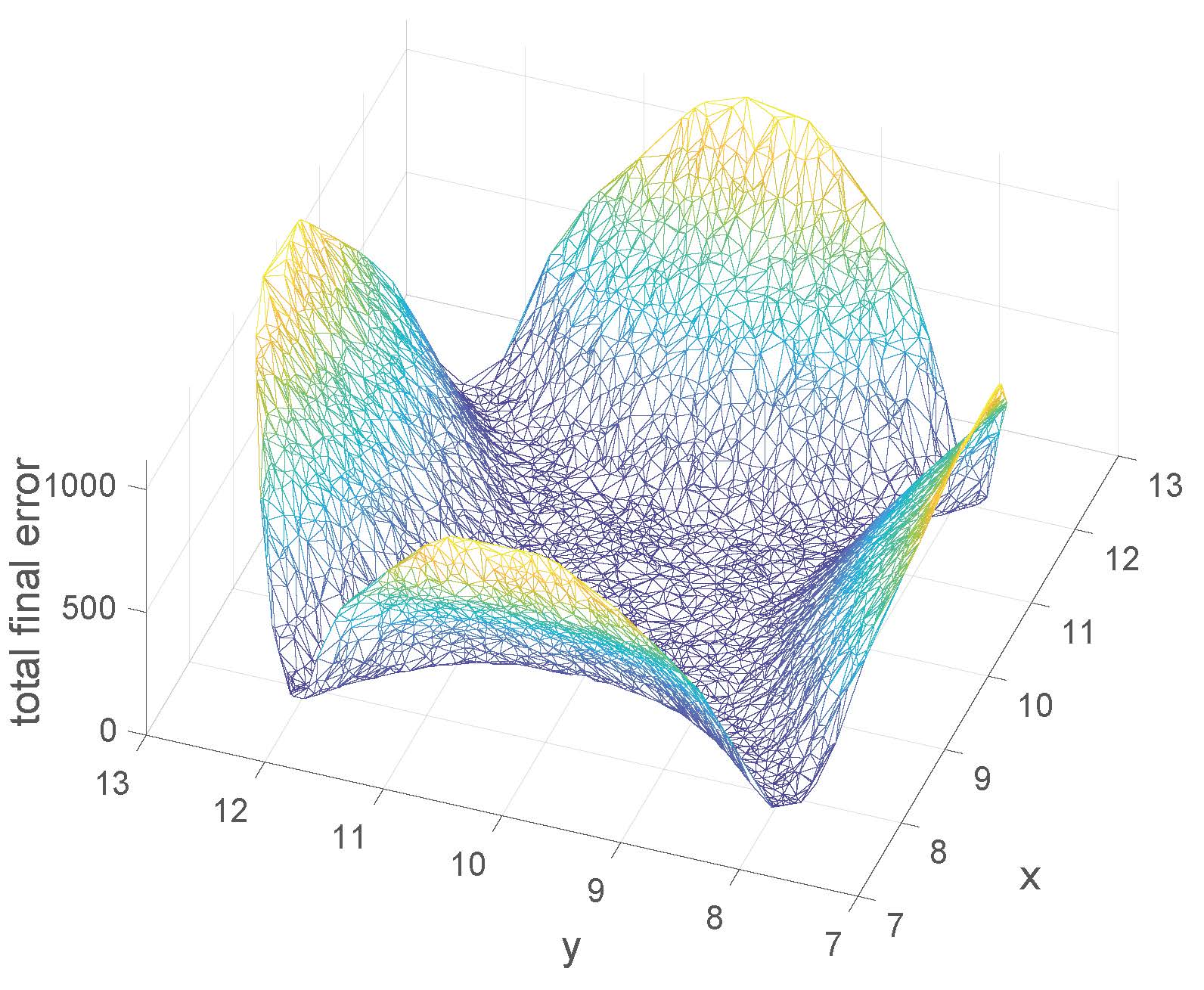}&
			\includegraphics[width=0.28\textwidth]{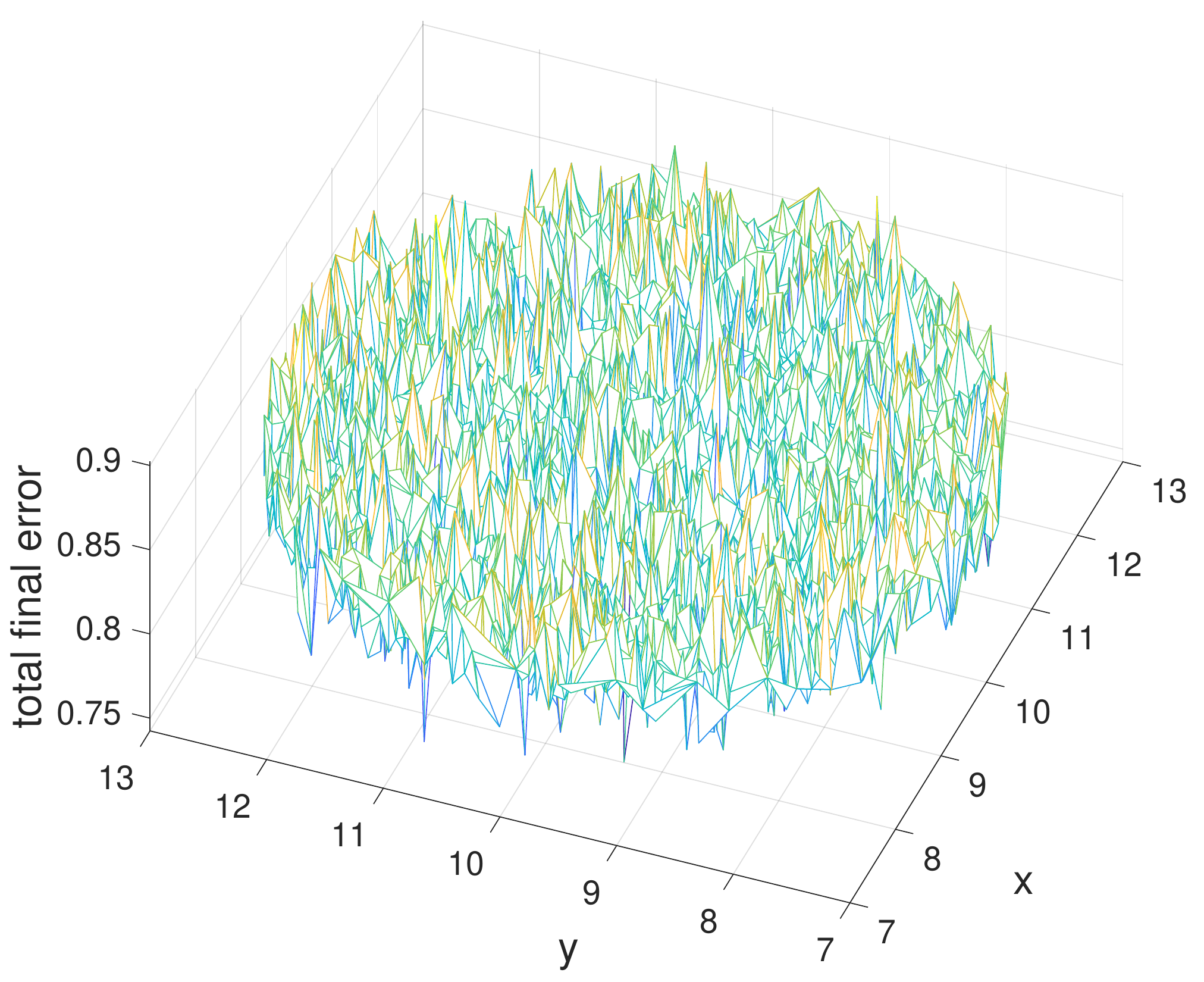}\\
			(a) $\mathcal{L}_{IoU}$ & (b) $\mathcal{L}_{GIoU}$ & (c) $\mathcal{L}_{DIoU}$
		\end{tabular}
		\caption{\footnotesize Visualization of regression errors of IoU, GIoU and DIoU losses at the final iteration $T$, \ie. $\mathbf{E}(T,n)$ for every coordinate $n$.
		We note that the basins in (a) and (b) correspond to good regression cases. One can see that IoU loss has large errors for non-overlapping cases, GIoU loss has large errors for horizontal and vertical cases, and our DIoU loss leads to very small regression errors everywhere.}
		\label{fig:finalerror}
	\end{figure*}

	In Fig. \ref{fig:finalerror}, we visualize the final regression errors at iteration $T$ for 5,000 scattered points.
	From Fig. \ref{fig:finalerror}(a), it is easy to see that IoU loss only works for the cases of overlapping with target boxes.
	The anchor boxes without overlap will not move due to that $\nabla B$ is always 0.
	
	By adding a penalty term as Eqn. \eqref{eq:giou loss}, GIoU loss can better relieve the issues of non-overlapping cases.
	From Fig. \ref{fig:finalerror}(b), GIoU loss significantly enlarges the basin, \ie, the area that GIoU works.
	But the cases at horizontal and vertical orientations are likely to still have large errors.
	This is because that the penalty term in GIoU loss is used to minimize $|C-A\cup B|$, but the area of $C-A\cup B$ is often small or 0 (when two boxes have inclusion relationships), and then GIoU almost degrades to IoU loss.
	%
    %
    GIoU loss would converge to good solution as long as running sufficient iterations with proper learning rates, but the convergence rate is indeed very slow.
	Geometrically speaking, from the regression steps as shown in Fig. \ref{fig:regression steps}, one can see that GIoU actually increases the predicted box size to overlap with target box, and then the IoU term will make the predicted box match with the target box, yielding a very slow convergence.

	To sum up, IoU loss converges to bad solutions for non-overlapping cases, while GIoU loss is with slow convergence especially for the boxes at horizontal and vertical orientations.
	And when incorporating into object detection pipeline, both IoU and GIoU losses cannot guarantee the accuracy of regression.
	It is natural to ask that:
	\emph{\textbf{First}}, is it feasible to directly minimize the normalized distance between predicted box and target box for achieving faster convergence?
	\emph{\textbf{Second}}, how to make the regression more accurate and faster when having overlap even inclusion with target box?

	\section{The Proposed Method}
	Generally, the IoU-based loss can be defined as
	\begin{equation}\label{eq:general loss}
	\mathcal{L}=1- IoU + \mathcal{R}(B,B^{gt}),
	\end{equation}
	where $\mathcal{R}(B,B^{gt})$ is the penalty term for predicted box $B$ and target box $B^{gt}$.
	By designing proper penalty terms, in this section we propose DIoU loss and CIoU loss to answer the aforementioned two questions.
	
	\subsection{Distance-IoU Loss}
\begin{figure}[!htb]
		\setlength{\abovecaptionskip}{0.cm}
		\setlength{\belowcaptionskip}{-0.cm}
		\centering
		\includegraphics[width=0.18\textwidth]{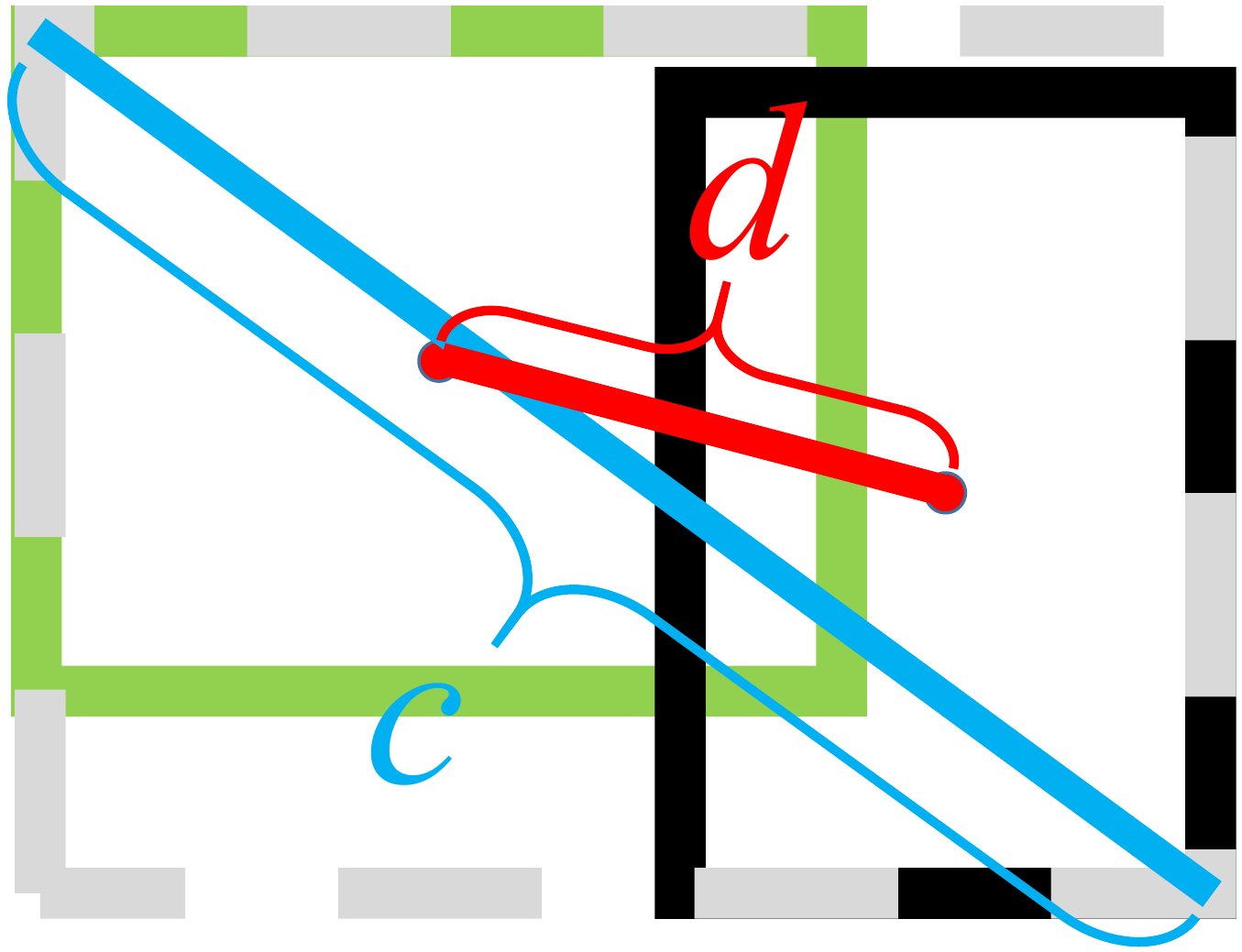}
		\caption{\footnotesize DIoU loss for bounding box regression, where the normalized distance between central points can be directly minimized.
		$c$ is the diagonal length of the smallest enclosing box covering two boxes, and $d=\rho(\mathbf{b},\mathbf{b}^{gt})$ is the distance of central points of two boxes. }
		\label{fig:DIoU}
	\end{figure}

	To answer the \textbf{\emph{first}} question, we propose to minimize the normalized distance between central points of two bounding boxes, and the penalty term can be defined as
	\begin{equation}\label{eq:penalty DIoU}
	\mathcal{R}_{DIoU} = \frac{\rho^2(\mathbf{b},\mathbf{b}^{gt})}{c^2},
	\end{equation}
	where $\mathbf{b}$ and $\mathbf{b}^{gt}$ denote the central points of $B$ and $B^{gt}$, $\rho(\cdot)$ is the Euclidean distance, and $c$ is the diagonal length of the smallest enclosing box covering the two boxes.
	And then the DIoU loss function can be defined as
	\begin{equation} \label{eq:riou loss}
	\mathcal{L}_{DIoU} = 1-IoU + \frac{\rho^2(\mathbf{b},\mathbf{b}^{gt})}{c^2}.
	\end{equation}
	As shown in Fig. \ref{fig:DIoU}, the penalty term of DIoU loss directly minimizes the distance between two central points, while GIoU loss aims to reduce the area of $C-B\cup B^{gt}$.

	\subsubsection{Comparison with IoU and GIoU losses}
	The proposed DIoU loss inherits some properties from IoU and GIoU loss.
	\begin{enumerate}
		
		\item
		DIoU loss is still invariant to the scale of regression problem.
		
		\item
		Similar to GIoU loss, DIoU loss can provide moving directions for bounding boxes when non-overlapping with target box.
		
		\item
		When two bounding boxes perfectly match, $\mathcal{L}_{IoU} = \mathcal{L}_{GIoU} = \mathcal{L}_{DIoU} = 0$.
		When two boxes are far away, $\mathcal{L}_{GIoU} = \mathcal{L}_{DIoU} \rightarrow 2$.

	\end{enumerate}

	And DIoU loss has several merits over IoU loss and GIoU loss, which can be evaluated by simulation experiment.
	\begin{enumerate}
		\item
		As shown in Fig. \ref{fig:regression steps} and Fig. \ref{fig:1715ksampling}, DIoU loss can directly minimize the distance of two boxes, and thus converges much faster than GIoU loss.
		
		\item
		For the cases with inclusion of two boxes, or in horizontal and vertical orientations, DIoU loss can make regression very fast, while GIoU loss has almost degraded to IoU loss, \ie, $|C-A\cup B| \rightarrow 0$.
		
	\end{enumerate}

	\subsection{Complete IoU Loss}
	
	Then we answer the \textbf{\emph{second}} question, by suggesting that a good loss for bounding box regression should consider three important geometric factors, \ie, overlap area, central point distance and aspect ratio.
	%
	By uniting the coordinates, IoU loss considers the overlap area, and GIoU loss heavily relies on IoU loss.
	Our proposed DIoU loss aims at considering simultaneously the overlap area and central point distance of bounding boxes.
	However, the consistency of aspect ratios for bounding boxes is also an important geometric factor.

	Therefore, based on DIoU loss, the CIoU loss is proposed by imposing the consistency of aspect ratio,
	\begin{equation}\label{eq:penalty complete}
	\mathcal{R}_{CIoU} = \frac{\rho^2(\mathbf{b},\mathbf{b}^{gt})}{c^2} + \alpha v,
	\end{equation}
	where $\alpha$ is a positive trade-off parameter, and $v$ measures the consistency of aspect ratio,
	\begin{equation}
	v =\frac{4}{\pi^{2}}(arctan\frac{w^{gt}}{h^{gt}}-arctan\frac{w}{h})^2.
	\end{equation}
	Then the loss function can be defined as
	\begin{equation}\label{eq:cDIoU loss}
	\mathcal{L}_{CIoU} = 1-IoU + \frac{\rho^2(\mathbf{b},\mathbf{b}^{gt})}{c^2} + \alpha v.
	\end{equation}
	%
	And the trade-off parameter $\alpha$ is defined as
	\begin{equation}\label{eq:alpha}
	\alpha=\frac{v}{(1-IoU)+ v},
	\end{equation}
	by which the overlap area factor is given higher priority for regression, especially for non-overlapping cases.

	Finally, the optimization of CIoU loss is same with that of DIoU loss, except that the gradient of $v$ \wrt $w$ and $h$ should be specified,
	\begin{equation}\label{eq:aspect gradient}\small
	\begin{aligned}
	\frac{\partial v}{\partial w}&=\frac{8}{\pi^2}(arctan\frac{w^{gt}}{h^{gt}}-arctan\frac{w}{h})\times \frac{h}{w^2+h^2},\\
	\frac{\partial v}{\partial h}&=-\frac{8}{\pi^2}(arctan\frac{w^{gt}}{h^{gt}}-arctan\frac{w}{h})\times \frac{w}{w^2+h^2}.
	\end{aligned}
	\end{equation}
	The dominator $w^{2}+h^2$ is usually a small value for the cases $h$ and $w$ ranging in $[0,1]$, which is likely to yield gradient explosion.
	And thus in our implementation, the dominator $w^{2}+h^2$ is simply removed for stable convergence, by which the step size $\frac{1}{w^{2}+h^2}$ is replaced by 1 and the gradient direction is still consistent with Eqn. \eqref{eq:aspect gradient}.

	\subsection{Non-Maximum Suppression using DIoU}
	
	In original NMS, the IoU metric is used to suppress the redundant detection boxes, where the overlap area is the unique factor, often yielding false suppression for the cases with occlusion.
	We in this work suggest that DIoU is a better criterion for NMS, because not only overlap area but also central point distance between two boxes should also be considered in the suppression criterion.
	For the predicted box $\mathcal{M}$ with the highest score, the DIoU-NMS can be formally defined as
	\begin{equation}
	{{s}_{i}}=\left\{
	\begin{aligned}
	& {{s}_{i}},\ IoU - \mathcal{R}_{DIoU}(\mathcal{M},B_i) < \varepsilon , \\
	& 0,\ \ IoU - \mathcal{R}_{DIoU}(\mathcal{M},B_i) \ge \varepsilon , \\
	\end{aligned} \right.
	\end{equation}
	where box $B_i$ is removed by simultaneously considering the IoU and the distance between central points of two boxes, $s_i$ is the classification score and $\varepsilon$ is the NMS threshold.
	We suggest that two boxes with distant central points probably locate different objects, and should not be removed.
	Moreover, the DIoU-NMS is very flexible to be integrated into any object detection pipeline with only a few lines of code.

	
	\section{Experimental Results}
	In this section, on two popular benchmarks including PASCAL VOC \cite{voc} and MS COCO \cite{coco}, we evaluate our proposed DIoU and CIoU losses by incorporating them into the state-of-the-art object detection algorithms including one-stage detection algorithms (\ie, YOLO v3 and SSD) and two-stage algorithm (\ie, Faster R-CNN).
	All the source codes and our trained models will be made publicly available.

	\subsection{YOLO v3 on PASCAL VOC}
	PASCAL VOC \cite{voc} is one of the most popular dataset for object detection.
	YOLO v3 is trained on PASCAL VOC using DIoU and CIoU losses in comparison with IoU and GIoU losses.
	We use VOC 07+12 (the union of VOC 2007 trainval and VOC 2012 trainval) as training set, containing 16,551 images from 20 classes.
	And the testing set is VOC 2007 test, which consists of 4,952 images.
	The backbone network is Darknet608.
	We follow exactly the GDarknet\footnote{\url{https://github.com/generalized-iou/g-darknet}} training protocol released from \cite{giou}, and the maximum iteration is set to 50K.
	The performance for each loss has been reported in Table~\ref{table:yolo-voc}.
	We use the same performance measure, \ie, AP (the average of 10 mAP across different IoU thresholds) = (AP50 + AP55 + $\dots$ + AP95) / 10 and AP75 (mAP@0.75).
	We also report the evaluation results using GIoU metric.
	
	\begin{table}[h]\footnotesize
		\setlength{\abovecaptionskip}{0.cm}
		\setlength{\belowcaptionskip}{-0.cm}
			\centering
			\caption{\footnotesize Quantitative comparison of \textbf{YOLOv3}~\cite{yolov3} trained using $\calL_{IoU}$ (baseline), $\calL_{GIoU}$, $\calL_{DIoU}$ and $\calL_{CIoU}$. (D) denotes using DIoU-NMS.
				The results are reported on the test set of PASCAL VOC 2007. }
			\begin{tabular}{c c c c c c}
				\hline
				\raisebox{-1.5ex}{Loss \big/ Evaluation} & \multicolumn{2}{c}{\raisebox{-0.5ex}{AP}} &&\multicolumn{2}{c}{\raisebox{-0.5ex}{AP75}}\\ [0.5ex]
				\cline{2-3}\cline{5-6}
				& \raisebox{-0.2ex}{\textbf{IoU}} & \raisebox{-0.2ex}{\textbf{GIoU}} &&\raisebox{-0.2ex}{\textbf{IoU}} & \raisebox{-0.2ex}{\textbf{GIoU}} \\
				\hline
				\hline
				$\calL_{IoU}$ & 46.57 & 45.82  && 49.82 & 48.76 \\
				\hline
				\hline
				$\calL_{GIoU}$ & 47.73 & 46.88 && 52.20 & 51.05 \\
				Relative improv. \% & 2.49\% & 2.31\% && 4.78\% & 4.70\% \\
				\hline
				$\calL_{DIoU}$ & 48.10 & 47.38 && 52.82 & 51.88 \\
				Relative improv. \% & 3.29\% & 3.40\% && 6.02\% & 6.40\% \\
				\hline
				$\calL_{CIoU}$ & \textbf{49.21} & \textbf{48.42} && \textbf{54.28} & \textbf{52.87} \\
				Relative improv. \% & \textbf{5.67\%} & \textbf{5.67\%} && \textbf{8.95\%} & \textbf{8.43\%} \\
				\hline
				$\calL_{CIoU}(D) $ & \textcolor{red}{\textbf{49.32}} & \textcolor{red}{\textbf{48.54}} && \textcolor{red}{\textbf{54.74}} & \textcolor{red}{\textbf{53.30}} \\
				Relative improv. \% & \textcolor{red}{\textbf{5.91\%}} & \textcolor{red}{\textbf{5.94\%}} && \textcolor{red}{\textbf{9.88\%}} & \textcolor{red}{\textbf{9.31\%}} \\
				\hline
			\end{tabular}
			\label{table:yolo-voc}
		
	\end{table}
	\begin{figure*}[!htb]\footnotesize
		\setlength{\abovecaptionskip}{0.cm}
		\setlength{\belowcaptionskip}{-0.cm}
	\centering
	\setlength{\tabcolsep}{1pt}
	\begin{tabular}{cccccccccc}
		\includegraphics[width=0.18\textwidth,height=2.3cm]{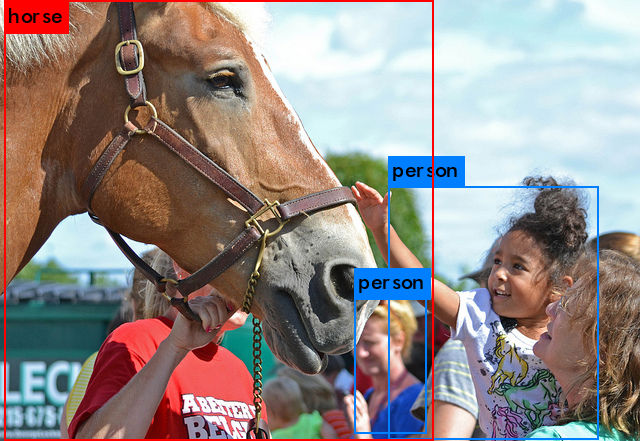}&
		\includegraphics[width=0.18\textwidth,height=2.3cm]{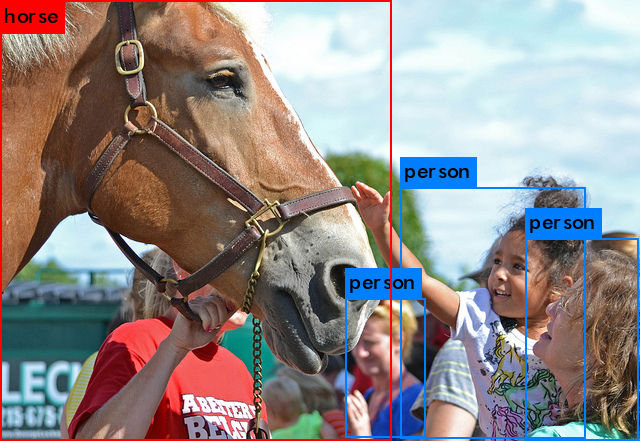}&
		\includegraphics[width=0.13\textwidth,height=2.3cm]{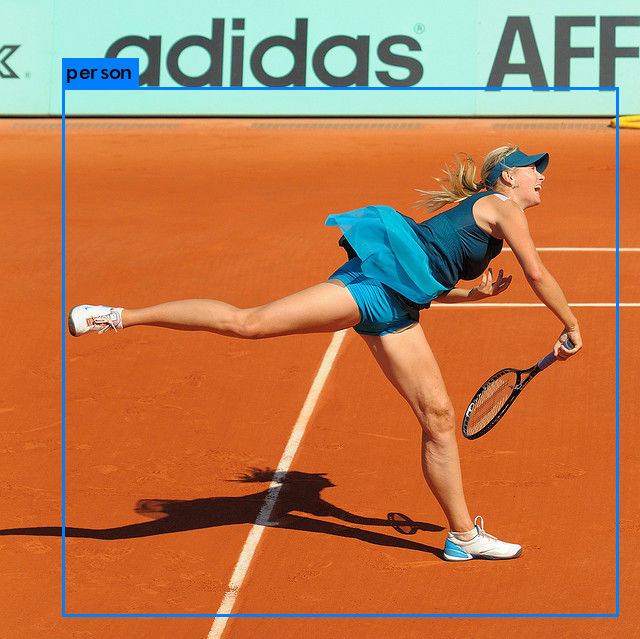}&
		\includegraphics[width=0.13\textwidth,height=2.3cm]{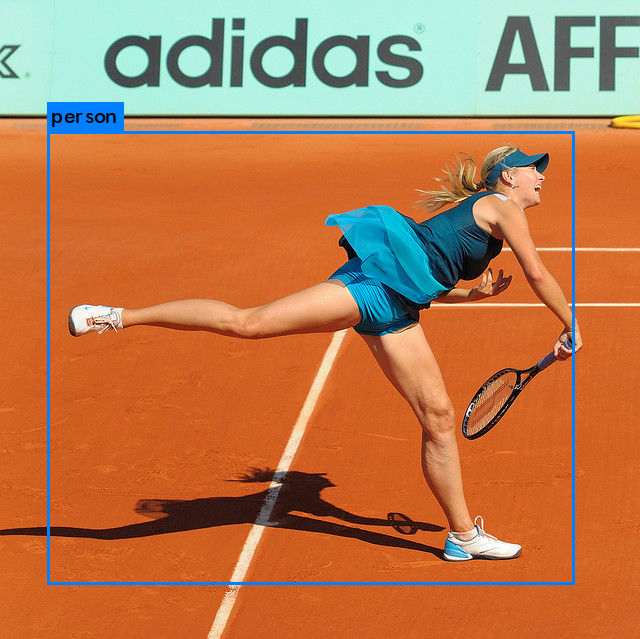}&
		\includegraphics[width=0.18\textwidth,height=2.3cm]{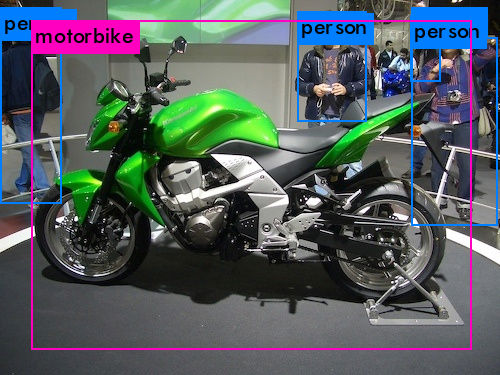}&
		\includegraphics[width=0.18\textwidth,height=2.3cm]{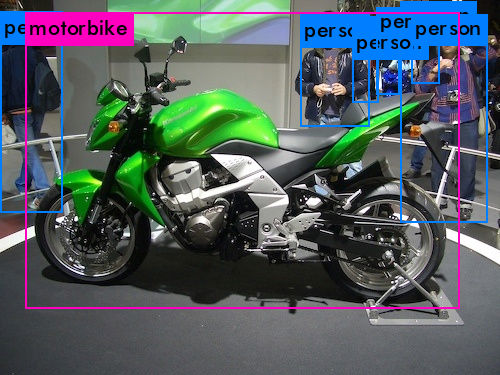}\\
		$\mathcal{L}_{GIoU}$ & $\mathcal{L}_{CIoU}$ & $\mathcal{L}_{GIoU}$  & $\mathcal{L}_{CIoU}$ & $\mathcal{L}_{GIoU}$  & $\mathcal{L}_{CIoU}$
	\end{tabular}
	\caption{Detection examples using YOLO v3 \cite{yolov3} trained on PASCAL VOC 07+12.}
	\label{fig:yolov3-examples}
\end{figure*}
	As shown in Table~\ref{table:yolo-voc}, GIoU as a generalized version of IoU, it indeed achieves a certain degree of performance improvement.
	While DIoU loss can improve the performance with gains of 3.29\% AP and 6.02\% AP75 using IoU as evaluation metric.
	CIoU loss takes the three important geometric factors of two bounding boxes into account, which brings an amazing performance gains, \ie, 5.67\% AP and 8.95\% AP75.
	From Fig. \ref{fig:yolov3-examples}, one can see that the detection box by CIoU loss is more accurate than that by GIoU loss.
	Finally, CIoU loss combined with DIoU-NMS brings marvelous improvements of 5.91\% AP and 9.88\% AP75.
	Also in terms of GIoU metric, we can come to the same conclusion, validating the effectiveness of the proposed methods.
	We note that GIoU metric is actually consistent with IoU metric, and thus we only report the IoU metric for the following experiments.
	
	\subsection{SSD on PASCAL VOC}
	We use another popular one-stage method SSD to further conduct evaluation experiments.
	The latest PyTorch implementation of SSD\footnote{\url{https://github.com/JaryHuang/awesome_SSD_FPN_GIoU}} is adopted.
	Both the training set and testing set share the same setting with YOLO v3 on PASCAL VOC.
	Following the default training protocol, the max iteration is set to 120K.
	The backbone network is ResNet-50-FPN.
	The default bounding box regression loss is smooth $\ell_1$-norm, which has different magnitudes with IoU-based losses.
	And thus there should be a more appropriate trade-off weight for the regression loss to balance with the classification loss.
	We have observed that for dense anchor algorithms, increasing the regression loss properly can improve the performance.
	Therefore, for a fair comparison,  we fix the weight on regression loss as 5 for these IoU-based losses.
	And then we train the models using IoU, GIoU, DIoU and CIoU losses.
	Table \ref{table:SSD-VOC} gives the quantitative comparison, in which AP and AP75 of IoU metric are reported.
	For SSD, we can see the consistent improvements of DIoU and CIoU losses in comparison with IoU and GIoU losses.
	\begin{table}[h]\footnotesize
		\setlength{\abovecaptionskip}{0.cm}
		\setlength{\belowcaptionskip}{-0.cm}
			\setlength{\tabcolsep}{15pt}
			\centering

			\caption{\footnotesize Quantitative comparison of \textbf{SSD}~\cite{SSD} trained using $\calL_{IoU}$ (baseline), $\calL_{GIoU}$, $\calL_{DIoU}$ and $\calL_{CIoU}$. (D) denotes using DIoU-NMS.
				The results are reported on the test set of PASCAL VOC 2007. }
			\begin{tabular}{c|ccc}
				\hline
				{Loss / Evaluation} & AP  & AP75 \\
				\hline
				\hline
				$\calL_{IoU}$& 51.01  & 54.74  \\
				\hline
				\hline
				$\calL_{GIoU}$& 51.06  & 55.48  \\
				Relative improv. \% & 0.10\%  & 1.35\% \\
				\hline
				$\calL_{DIoU}$ & 51.31  & 55.71  \\
				Relative improv. \% & 0.59\%  & 1.77\%  \\
				\hline
				$\calL_{CIoU}$ & \textbf{51.44}  & \textbf{56.16}  \\
				Relative improv. \% & \textbf{0.84\%}  & \textbf{2.59\%} \\
				\hline
				$\calL_{CIoU}(D)$  & \textcolor{red}{\textbf{51.63}} & \textcolor{red}{\textbf{56.34}}  \\
				Relative improv. \% & \textcolor{red}{\textbf{1.22\%}}  & \textcolor{red}{\textbf{2.92\%}} \\
				\hline
			\end{tabular}
			\label{table:SSD-VOC}
		
	\end{table}
	
		\begin{figure*}[!tb]\footnotesize
			\setlength{\abovecaptionskip}{0.cm}
			\setlength{\belowcaptionskip}{-0.cm}
		\centering
		\setlength{\tabcolsep}{1pt}
		\begin{tabular}{cccccccccc}

			\includegraphics[width=0.17\textwidth,height=2cm]{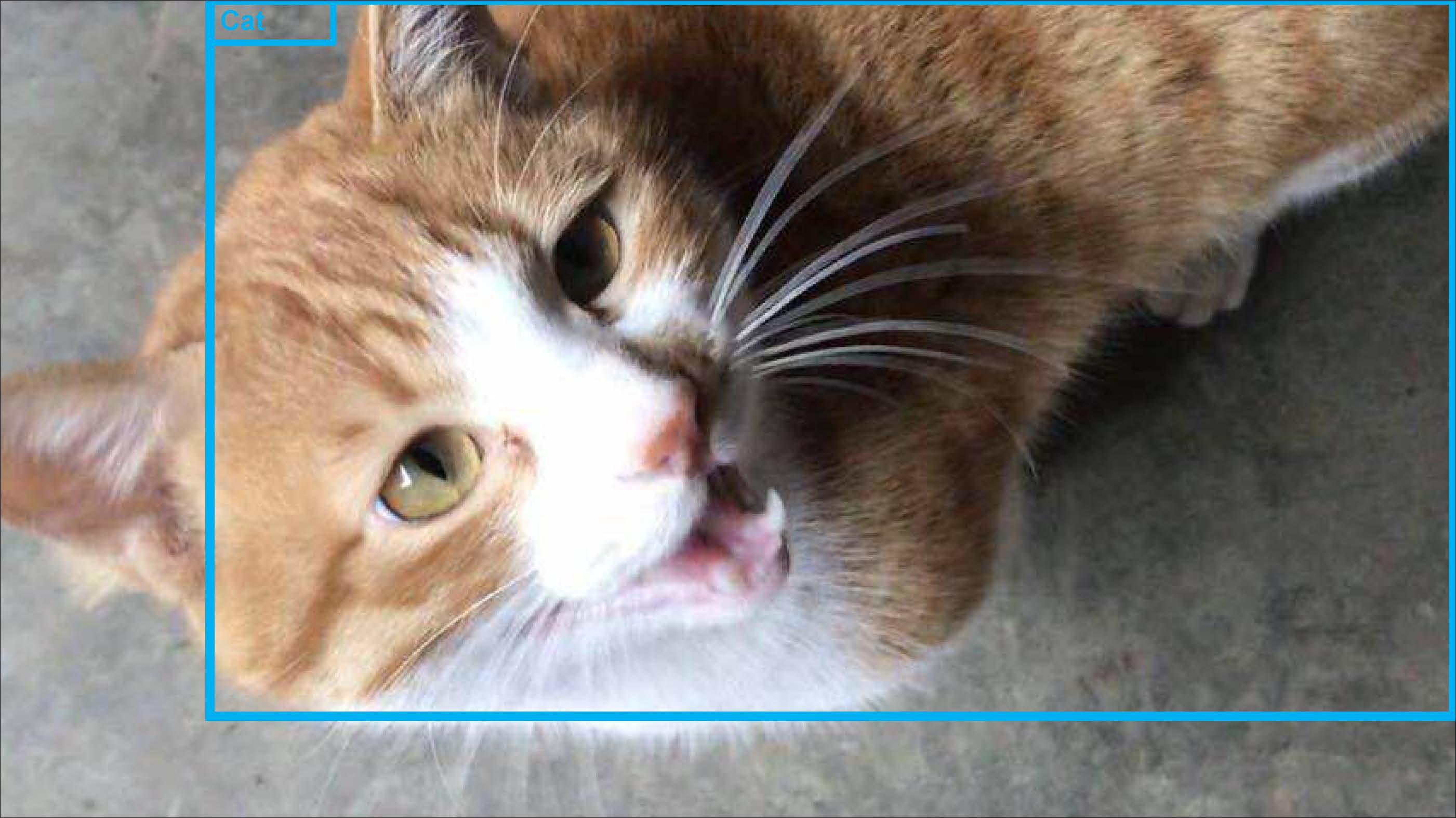}&
			\includegraphics[width=0.17\textwidth,height=2cm]{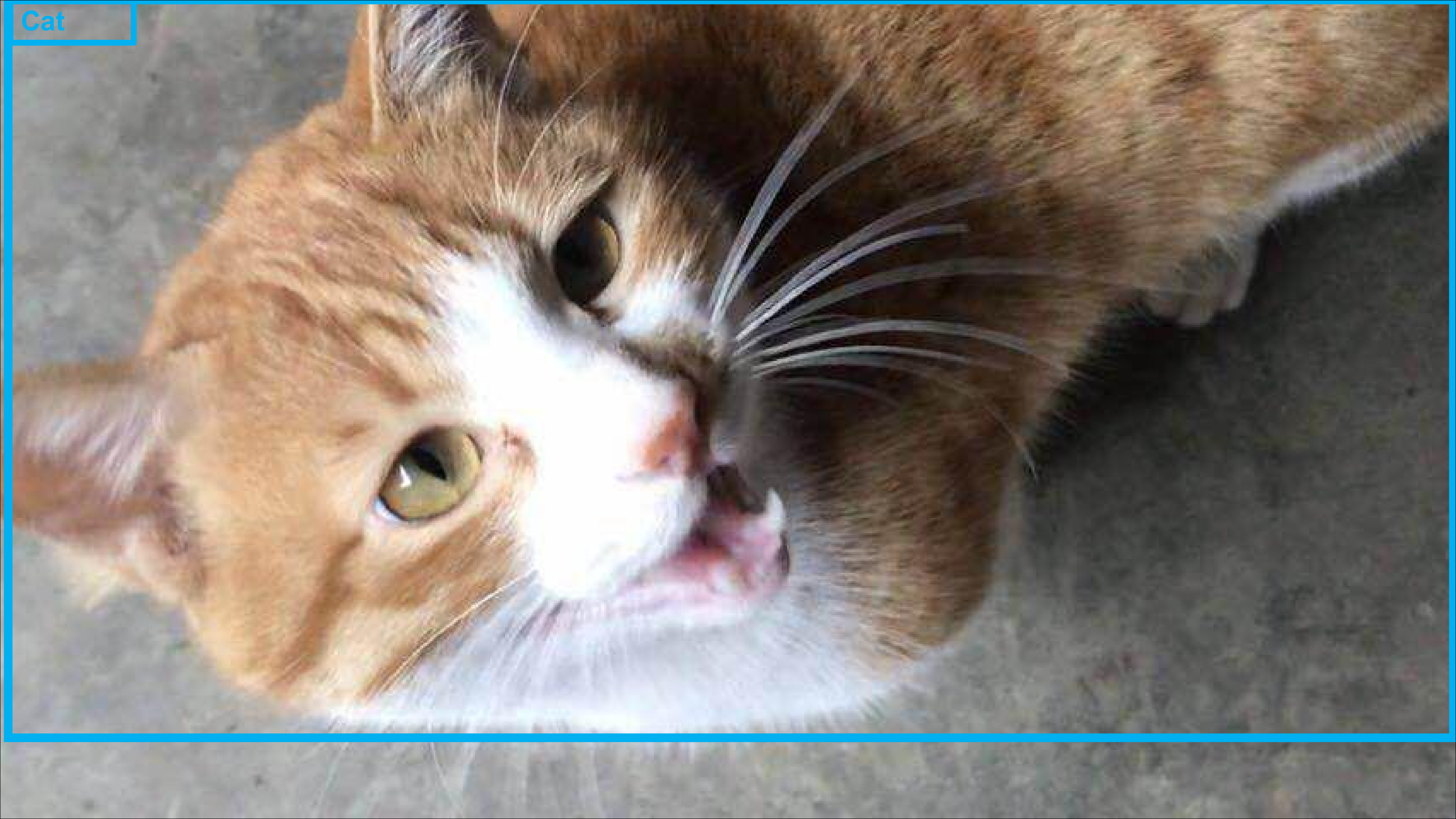}&
			\includegraphics[width=0.16\textwidth,height=2cm]{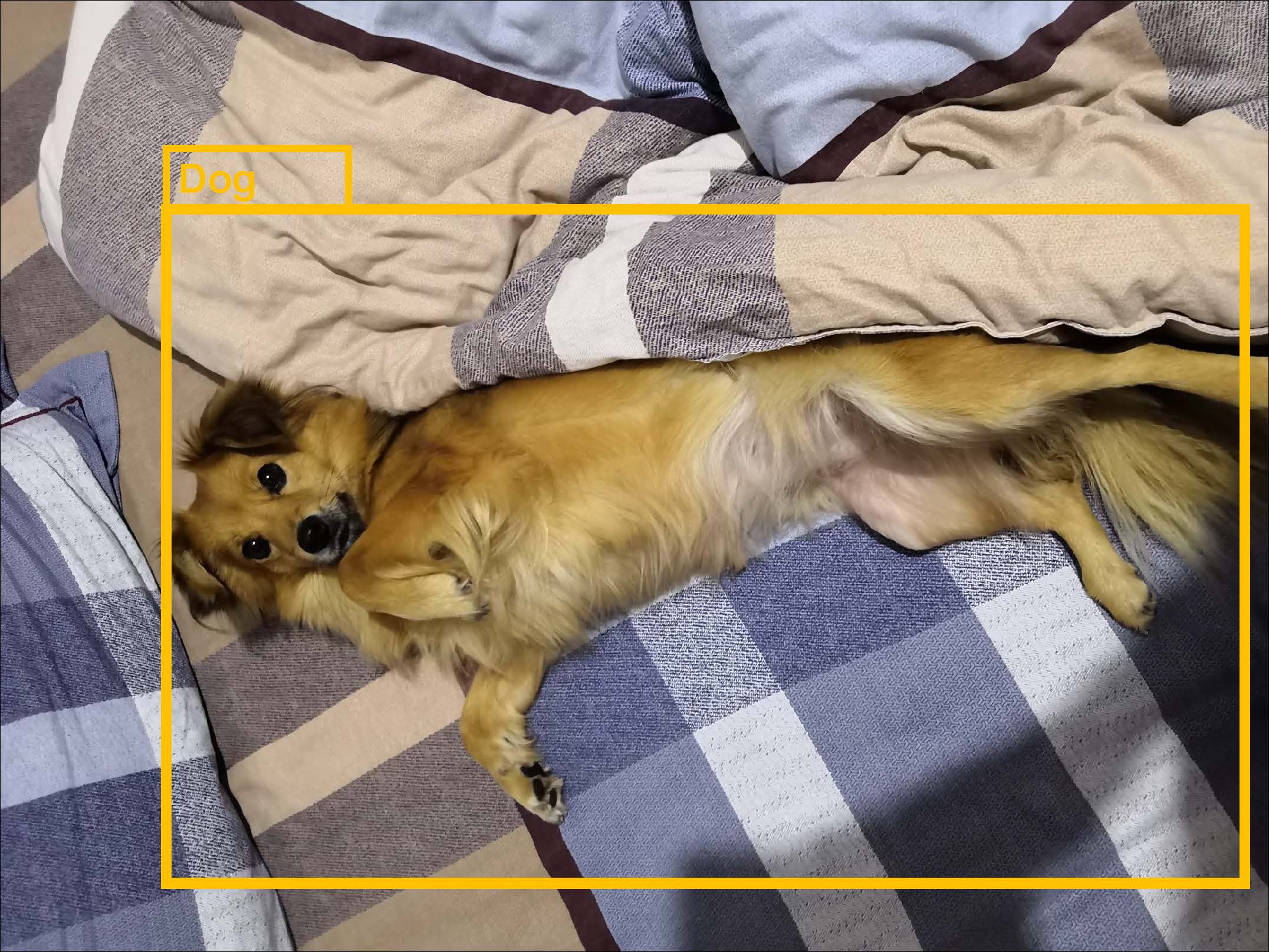}&
			\includegraphics[width=0.16\textwidth,height=2cm]{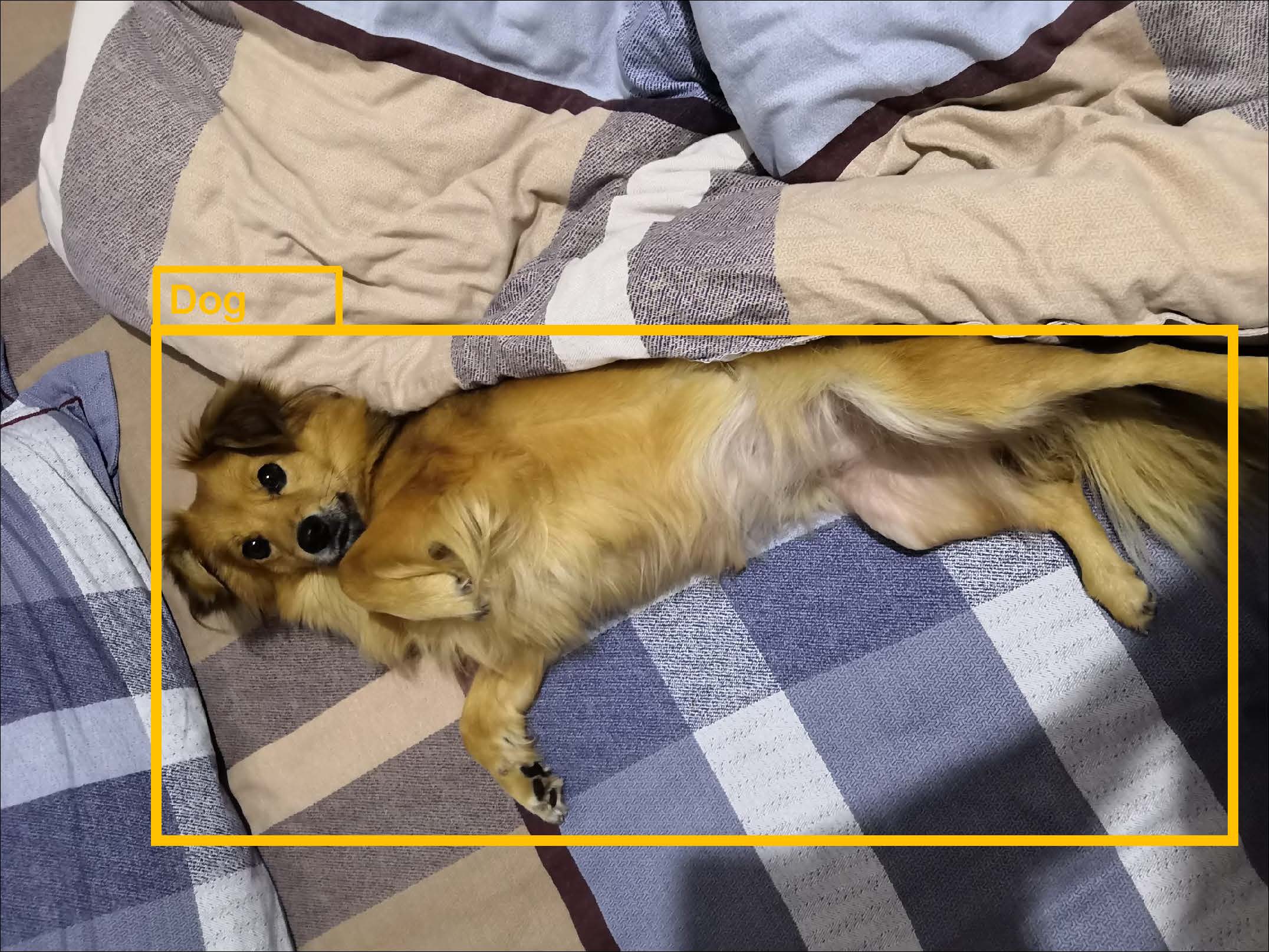}&
			\includegraphics[width=0.16\textwidth,height=2cm]{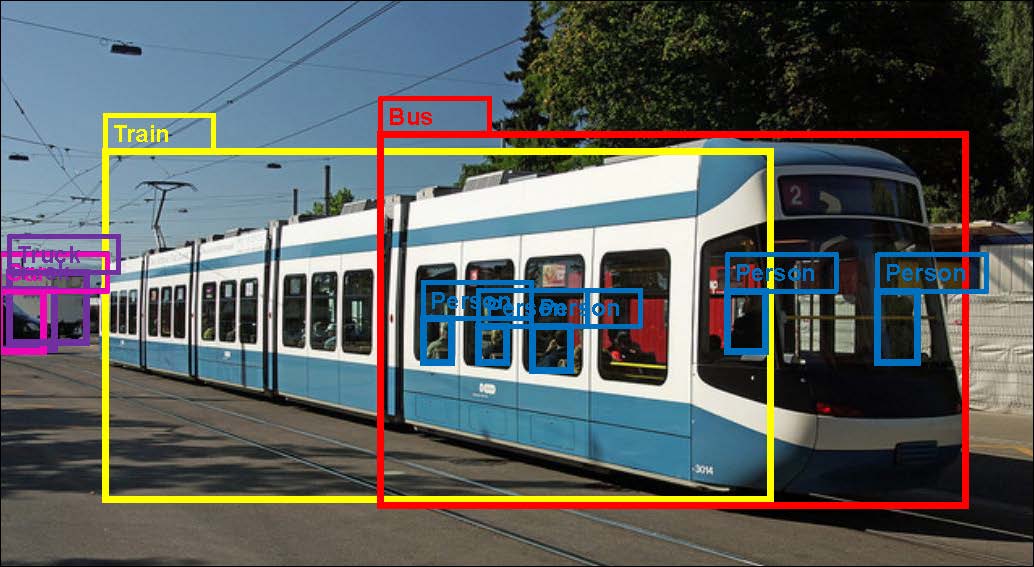}&
			\includegraphics[width=0.16\textwidth,height=2cm]{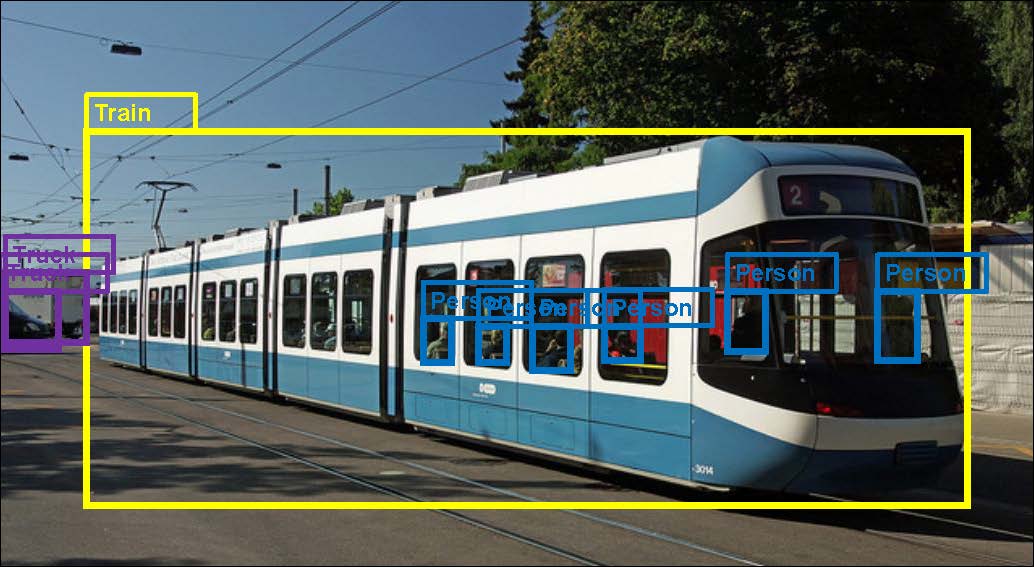}\\
			$\mathcal{L}_{GIoU}$ & $\mathcal{L}_{CIoU}$ & $\mathcal{L}_{GIoU}$  & $\mathcal{L}_{CIoU}$ & $\mathcal{L}_{GIoU}$  & $\mathcal{L}_{CIoU}$
		\end{tabular}
		\caption{\footnotesize Detection examples using Faster R-CNN \cite{fasterrcnn} trained on MS COCO 2017.}
		\label{fig:fasterrcnn-examples}
	\end{figure*}

	\subsection{Faster R-CNN on MS COCO}
	\begin{table}[!htb]
		\footnotesize
		\setlength{\abovecaptionskip}{0.cm}
		\setlength{\belowcaptionskip}{-0.cm}
			\setlength{\tabcolsep}{2pt}
			\centering
			\caption{\footnotesize Quantitative comparison of \textbf{Faster R-CNN}~\cite{fasterrcnn} trained using $\calL_{IoU}$ (baseline), $\calL_{GIoU}$, $\calL_{DIoU}$ and $\calL_{CIoU}$. (D) denotes using DIoU-NMS.
				The results are reported on the validation set of MS COCO 2017.}
			\begin{tabular}{c|cc|ccc}
				\hline
				\multirow{2}{*}{Loss / Evaluation} & \multirow{2}{*}{AP}  & \multirow{2}{*}{AP75}& \multirow{2}{*}{AP\small{small}} & \multirow{2}{*}{AP\small{medium}} & \multirow{2}{*}{AP\small{large}}  \\
				&&&&&  \\
				\hline
				\hline
				$\calL_{IoU}$& 37.93  & 40.79 & 21.58 & 40.82 & 50.14 \\
				\hline
				\hline
				$\calL_{GIoU}$& 38.02 & 41.11 & 21.45 & 41.06 & 50.21 \\
				Relative improv. \% & 0.24\%  & 0.78\% & \text{-}0.60\% & 0.59\% & 0.14\% \\
				\hline
				$\calL_{DIoU}$ & 38.09  & 41.11 & \textbf{21.66} & 41.18 & 50.32 \\
				Relative improv. \% & 0.42\%  & 0.78\% & \textbf{0.31\%} & 0.88\% & 0.36\% \\
				\hline
				$\calL_{CIoU}$ & \textbf{38.65}  & \textbf{41.96} & 21.32 & \textbf{41.83} & \textbf{51.51} \\
				Relative improv. \% & \textbf{1.90\%}  & \textbf{2.87\%} & \text{-}1.20\% & \textbf{2.47\%} & \textbf{2.73\%} \\
				\hline
				$\calL_{CIoU}(D)$ & \textcolor{red}{\textbf{38.71}}  & \textcolor{red}{\textbf{42.07}} & 21.37 & \textcolor{red}{\textbf{41.93}} & \textcolor{red}{\textbf{51.60}} \\
				Relative improv. \% & \textcolor{red}{\textbf{2.06\%}} &  \textcolor{red}{\textbf{3.14\%}} & \text{-}0.97\% & \textcolor{red}{\textbf{2.72\%}} & \textcolor{red}{\textbf{2.91\%}} \\
				\hline
			\end{tabular}
			\label{table:faster-rcnn-cocoval}
		
	\end{table}
	We also evaluate the proposed method on another more difficult and complex dataset MS COCO 2017 \cite{coco} using Faster R-CNN\footnote{\url{https://github.com/generalized-iou/Detectron.pytorch}}.
	MS COCO is a large-scale dataset, containing more than 118K images for training and 5K images for evaluation.
	Following the same training protocol of \cite{giou}, we trained the models using DIoU and CIoU losses in comparison with IoU and GIoU losses.
	The backbone network is ResNet-50-FPN.
	Besides AP and AP75 metrics, the evaluation metrics in terms of large, medium and small scale objects are also included.
	As for the trade-off weight for regression loss, we set the weight as 12 for all losses for a fair comparison.
	\Tab~\ref{table:faster-rcnn-cocoval} reports the quantitative comparison.
	
	Faster R-CNN is a detection algorithm with dense anchor boxes, and is usually with high IoU levels in the initial situation.
	Geometrically speaking, the regression cases of Faster R-CNN are likely to place in the basins of Fig. \ref{fig:finalerror}, where IoU, GIoU and DIoU losses all have good performance.
	Therefore, GIoU loss has very small gain than the baseline IoU loss, as shown in \Tab~\ref{table:faster-rcnn-cocoval}.
	But our DIoU and CIoU losses still contribute to performance improvements than IoU and GIoU losses in terms of AP, AP75, APmedium and APlarge.
	Especially the gains by CIoU loss are very significant.
	From Fig. \ref{fig:fasterrcnn-examples}, one can easily find more accurate detection boxes by CIoU loss than those by GIoU loss.
	One may have noticed that in terms of APsmall, CIoU loss is a little inferior to the original IoU loss, while DIoU loss is better than all the other losses.
	That is to say the consistency of aspect ratio may not contribute to the regression accuracy for small objects.
	Actually it is reasonable that for small objects, the central point distance is more important than aspect ratio for regression, and the aspect ratio may weaken the effect of normalized distance between the two boxes.
	Nevertheless, CIoU loss performs much better for medium and large objects, and for small objects, the adverse effects can be relieved by DIoU-NMS.
	
	\subsection{Discussion on DIoU-NMS}

\begin{figure}[!htb]\footnotesize
		\setlength{\abovecaptionskip}{0.cm}
		\setlength{\belowcaptionskip}{-0.cm}
	\centering
	\setlength{\tabcolsep}{1pt}
	\begin{tabular}{cccccccccc}
		\includegraphics[height=3cm]{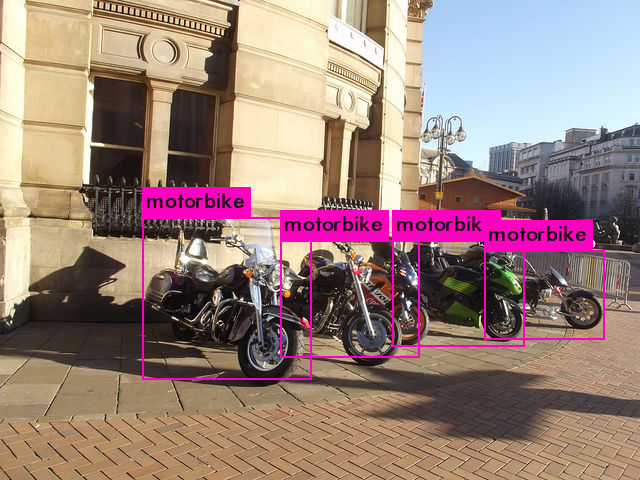}&
		\includegraphics[height=3cm]{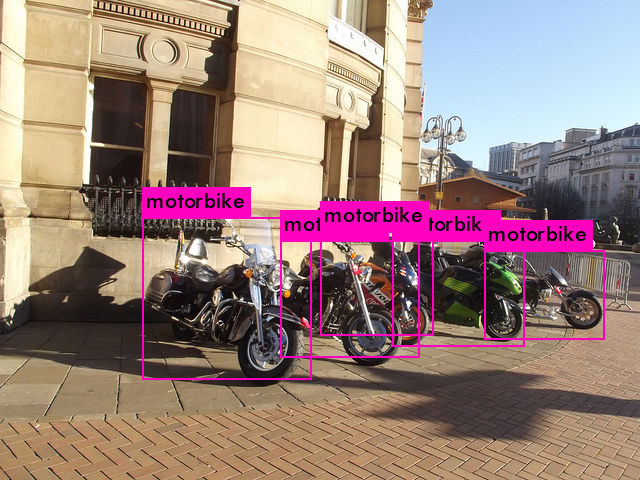}\\
		$\mathcal{L}_{CIoU}$ + NMS  & $\mathcal{L}_{CIoU}$ + DIoU-NMS
	\end{tabular}
	\caption{\footnotesize Detection example from MS COCO 2017 using YOLO v3 \cite{yolov3} trained on PASCAL VOC 07+12. }
	\label{fig:originalnms-vs-DIoUnms}
\end{figure}

\begin{figure}[!htb]\footnotesize
	\setlength{\abovecaptionskip}{0.cm}
	\setlength{\belowcaptionskip}{-0.cm}
	\centering
	\setlength{\tabcolsep}{0pt}
	\begin{tabular}{cccccccccc}
		\includegraphics[width=0.42\textwidth]{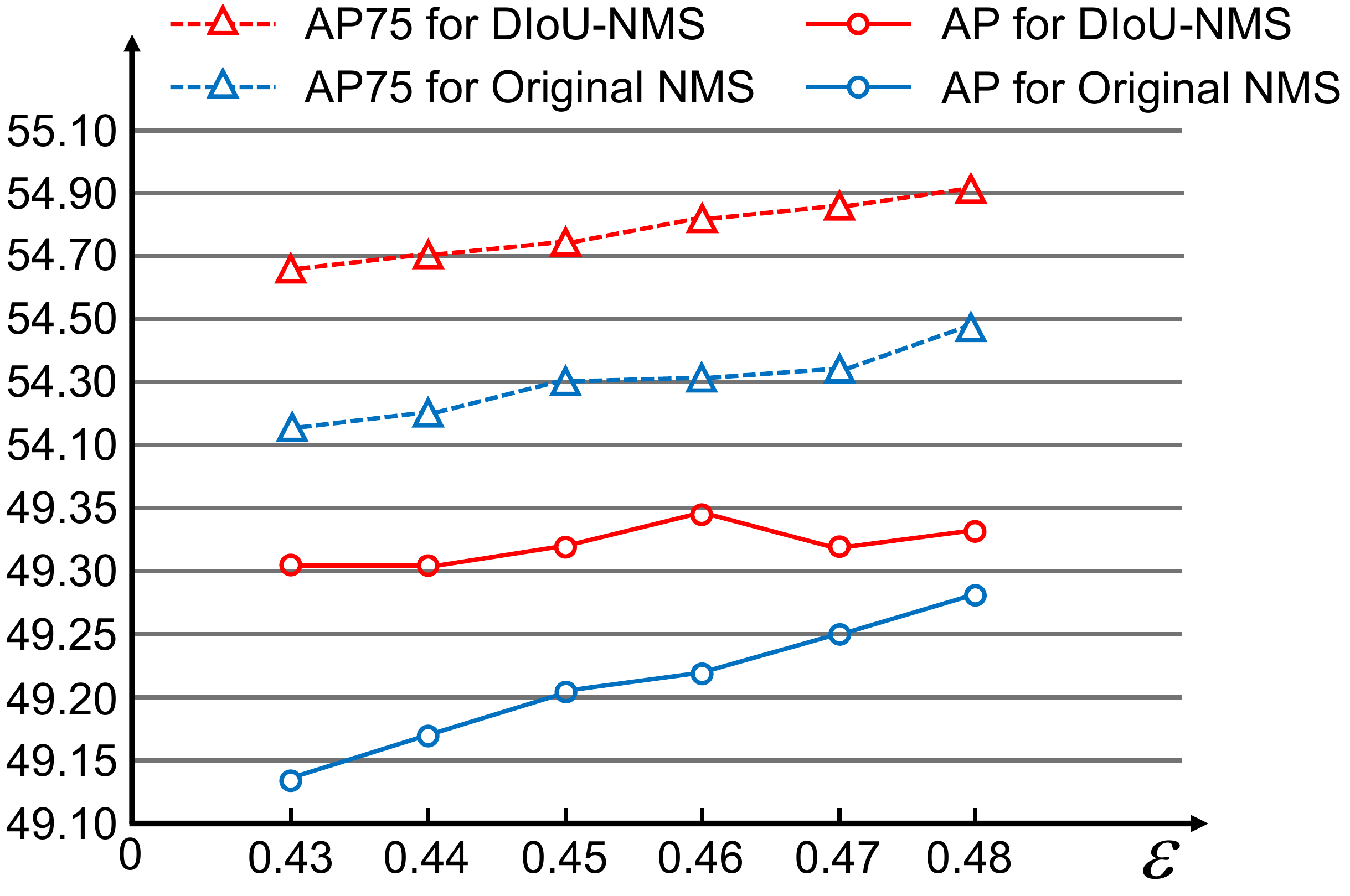}\\
		(a) YOLO v3 on the test set of PASCAL VOC 2007\\
		\includegraphics[width=0.42\textwidth]{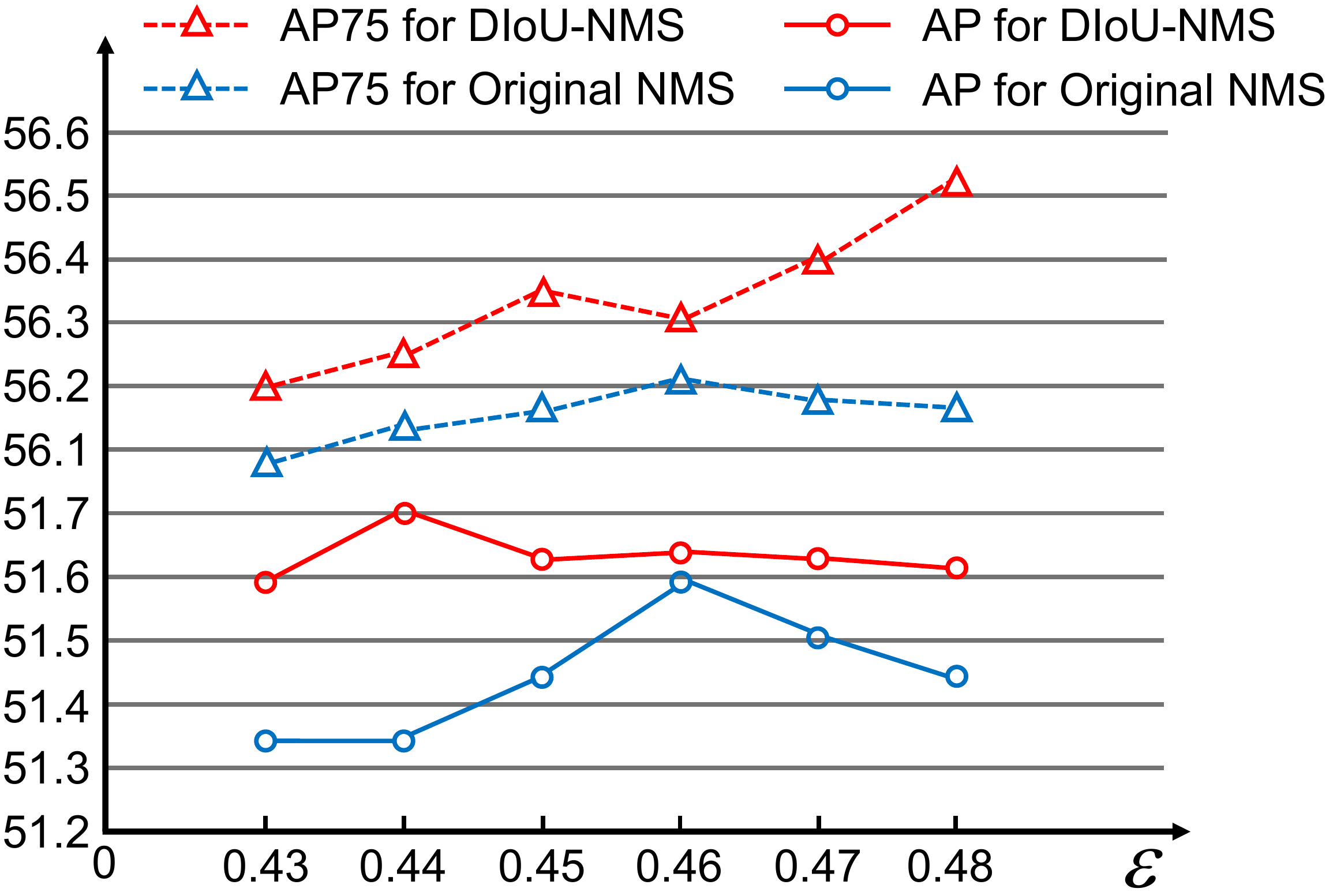}\\
		(b) SSD on the test set of PASCAL VOC 2007\\
	\end{tabular}
	\caption{\footnotesize Comparison of DIoU-NMS and original NMS for different thresholds $\varepsilon$. The models of YOLO v3 and SSD are trained on PASCAL VOC 07+12 using $\mathcal{L}_{CIoU}$.}
	\label{fig:DIoU-nms}
\end{figure}

	In Tables~\ref{table:yolo-voc}, \ref{table:SSD-VOC} and \ref{table:faster-rcnn-cocoval}, we report the results of CIoU loss cooperating with original NMS ($\mathcal{L}_{CIoU}$) and DIoU-NMS ($\mathcal{L}_{CIoU}(D)$), where the thresholds follow the default settings of original NMS, \ie, $\varepsilon=0.45$ for YOLO v3 and SSD, and $\varepsilon=0.50$ for Faster R-CNN.
	One can find that DIoU-NMS makes further performance improvements than original NMS for most cases.
	Fig. \ref{fig:originalnms-vs-DIoUnms} shows that DIoU-NMS can better preserve the correct detection boxes, where YOLO v3 trained on PASCAL VOC is adopted to detect objects on MS COCO.
	To further validate the superiority of DIoU-NMS over original NMS, we conduct comparison experiments, where original NMS and DIoU-NMS are cooperated with YOLO v3 and SSD trained using CIoU loss.
	We present the comparison of original NMS and DIoU-NMS within a wide range of thresholds $[0.43, 0.48]$.
	From Fig. \ref{fig:DIoU-nms}, one can see that DIoU-NMS is better than original NMS for every threshold.
	Furthermore, it is worth noting that even the worst performance of DIoU-NMS is at least comparable or better than the best performance of original NMS.
	That is to say our DIoU-NMS can generally perform better than original NMS even without carefully tuning the threshold $\varepsilon$.	

	
	\section{Conclusion}

	In this paper, we proposed two losses, \ie, DIoU loss and CIoU loss, for bounding box regression along with DIoU-NMS for suppressing redundant detection boxes.
	By directly minimizing the normalized distance of two central points, DIoU loss can achieve faster convergence than GIoU loss.
	CIoU loss takes three geometric properties into account, \ie, overlap area, central point distance and aspect ratio, and leads to faster convergence and better performance.
	The proposed losses and DIoU-NMS can be easily incorporated to any object detection pipeline, and achieve superior results on benchmarks.
	
	\section{Acknowledgments}
	This work is supported in part by the National Natural Science Foundation of China (Nos. 91746107 and 61801326) and the Operating Expenses of Basic Scientific Research Projects of the People's Public Security University of China Grant (Nos.2018JKF617 and 2019JKF111).
	We also thank Prof. Wangmeng Zuo, Zhanjie Song and Jun Wang for their valuable suggestions and favors.
	
	{\small
		\bibliographystyle{aaai}
		\bibliography{dioucite}

\begin{thebibliography}{}

\bibitem[\protect\citeauthoryear{Bae}{2019}]{RDAD}
Bae, S.-H.
\newblock 2019.
\newblock Object detection based on region decomposition and assembly.
\newblock In {\em The AAAI Conference on Artificial Intelligence}.

\bibitem[\protect\citeauthoryear{Bodla \bgroup et al\mbox.\egroup
  }{2017}]{softnms}
Bodla, N.; Singh, B.; Chellappa, R.; and Davis, L.~S.
\newblock 2017.
\newblock Soft-nms -- improving object detection with one line of code.
\newblock In {\em The IEEE International Conference on Computer Vision (ICCV)}.

\bibitem[\protect\citeauthoryear{Cai and Vasconcelos}{2018}]{cascadercnn}
Cai, Z., and Vasconcelos, N.
\newblock 2018.
\newblock Cascade r-cnn: Delving into high quality object detection.
\newblock In {\em The IEEE Conference on Computer Vision and Pattern
  Recognition (CVPR)}.

\bibitem[\protect\citeauthoryear{Cui \bgroup et al\mbox.\egroup
  }{2019}]{classbalancedloss}
Cui, Y.; Jia, M.; Lin, T.-Y.; Song, Y.; and Belongie, S.
\newblock 2019.
\newblock Class-balanced loss based on effective number of samples.
\newblock In {\em The IEEE Conference on Computer Vision and Pattern
  Recognition (CVPR)}.

\bibitem[\protect\citeauthoryear{Everingham \bgroup et al\mbox.\egroup
  }{2010}]{voc}
Everingham, M.; Van~Gool, L.; Williams, C. K.~I.; Winn, J.; and Zisserman, A.
\newblock 2010.
\newblock The pascal visual object classes (voc) challenge.
\newblock {\em International Journal of Computer Vision} 88(2):303--338.

\bibitem[\protect\citeauthoryear{Fu \bgroup et al\mbox.\egroup }{2017}]{DSSD}
Fu, C.-Y.; Liu, W.; Ranga, A.; Tyagi, A.; and Berg, A.~C.
\newblock 2017.
\newblock {DSSD}: Deconvolutional single shot detector.
\newblock {\em arXiv:1701.06659}.

\bibitem[\protect\citeauthoryear{Girshick \bgroup et al\mbox.\egroup
  }{2014}]{rcnn}
Girshick, R.; Donahue, J.; Darrell, T.; and Malik, J.
\newblock 2014.
\newblock Rich feature hierarchies for accurate object detection and semantic
  segmentation.
\newblock In {\em The IEEE Conference on Computer Vision and Pattern
  Recognition (CVPR)}.

\bibitem[\protect\citeauthoryear{Girshick}{2015}]{fastrcnn}
Girshick, R.
\newblock 2015.
\newblock Fast r-cnn.
\newblock In {\em The IEEE International Conference on Computer Vision (ICCV)}.

\bibitem[\protect\citeauthoryear{He \bgroup et al\mbox.\egroup
  }{2017}]{maskrcnn}
He, K.; Gkioxari, G.; Dollar, P.; and Girshick, R.
\newblock 2017.
\newblock Mask r-cnn.
\newblock In {\em The IEEE International Conference on Computer Vision (ICCV)}.

\bibitem[\protect\citeauthoryear{He \bgroup et al\mbox.\egroup
  }{2019}]{softernms}
He, Y.; Zhu, C.; Wang, J.; Savvides, M.; and Zhang, X.
\newblock 2019.
\newblock Bounding box regression with uncertainty for accurate object
  detection.
\newblock In {\em The IEEE Conference on Computer Vision and Pattern
  Recognition (CVPR)}.

\bibitem[\protect\citeauthoryear{Jiang \bgroup et al\mbox.\egroup
  }{2018}]{iounet}
Jiang, B.; Luo, R.; Mao, J.; Xiao, T.; and Jiang, Y.
\newblock 2018.
\newblock Acquisition of localization confidence for accurate object detection.
\newblock In {\em The European Conference on Computer Vision (ECCV)}.

\bibitem[\protect\citeauthoryear{Law and Deng}{2018}]{cornernet}
Law, H., and Deng, J.
\newblock 2018.
\newblock Cornernet: Detecting objects as paired keypoints.
\newblock In {\em The European Conference on Computer Vision (ECCV)}.

\bibitem[\protect\citeauthoryear{Li, Liu, and Wang}{2019}]{GHM}
Li, B.; Liu, Y.; and Wang, X.
\newblock 2019.
\newblock Gradient harmonized single-stage detector.
\newblock In {\em The AAAI Conference on Artificial Intelligence}.

\bibitem[\protect\citeauthoryear{Lin \bgroup et al\mbox.\egroup }{2014}]{coco}
Lin, T.-Y.; Maire, M.; Belongie, S.; Hays, J.; Perona, P.; Ramanan, D.;
  Doll{\'a}r, P.; and Zitnick, C.~L.
\newblock 2014.
\newblock Microsoft coco: Common objects in context.
\newblock In {\em The European Conference on Computer Vision (ECCV)}.

\bibitem[\protect\citeauthoryear{Lin \bgroup et al\mbox.\egroup
  }{2017}]{focalloss}
Lin, T.-Y.; Goyal, P.; Girshick, R.; He, K.; and Dollar, P.
\newblock 2017.
\newblock Focal loss for dense object detection.
\newblock In {\em The IEEE International Conference on Computer Vision (ICCV)}.

\bibitem[\protect\citeauthoryear{Liu \bgroup et al\mbox.\egroup }{2016}]{SSD}
Liu, W.; Anguelov, D.; Erhan, D.; Szegedy, C.; Reed, S.; Fu, C.-Y.; and Berg,
  A.~C.
\newblock 2016.
\newblock Ssd: Single shot multibox detector.
\newblock In {\em The European Conference on Computer Vision (ECCV)}.

\bibitem[\protect\citeauthoryear{Liu, Huang, and Wang}{2019}]{adaptivenms}
Liu, S.; Huang, D.; and Wang, Y.
\newblock 2019.
\newblock Adaptive nms: Refining pedestrian detection in a crowd.
\newblock In {\em The IEEE Conference on Computer Vision and Pattern
  Recognition (CVPR)}.

\bibitem[\protect\citeauthoryear{Pang \bgroup et al\mbox.\egroup
  }{2019}]{librarcnn}
Pang, J.; Chen, K.; Shi, J.; Feng, H.; Ouyang, W.; and Lin, D.
\newblock 2019.
\newblock Libra r-cnn: Towards balanced learning for object detection.
\newblock In {\em The IEEE Conference on Computer Vision and Pattern
  Recognition (CVPR)}.

\bibitem[\protect\citeauthoryear{Redmon and Farhadi}{2017}]{yolov2}
Redmon, J., and Farhadi, A.
\newblock 2017.
\newblock Yolo9000: Better, faster, stronger.
\newblock In {\em The IEEE Conference on Computer Vision and Pattern
  Recognition (CVPR)}.

\bibitem[\protect\citeauthoryear{Redmon and Farhadi}{2018}]{yolov3}
Redmon, J., and Farhadi, A.
\newblock 2018.
\newblock Yolov3: An incremental improvement.
\newblock {\em arXiv:1804.02767}.

\bibitem[\protect\citeauthoryear{Redmon \bgroup et al\mbox.\egroup
  }{2016}]{yolov1}
Redmon, J.; Divvala, S.; Girshick, R.; and Farhadi, A.
\newblock 2016.
\newblock You only look once: Unified, real-time object detection.
\newblock In {\em The IEEE Conference on Computer Vision and Pattern
  Recognition (CVPR)}.

\bibitem[\protect\citeauthoryear{Ren \bgroup et al\mbox.\egroup
  }{2015}]{fasterrcnn}
Ren, S.; He, K.; Girshick, R.; and Sun, J.
\newblock 2015.
\newblock Faster r-cnn: Towards real-time object detection with region proposal
  networks.
\newblock In {\em Advances in Neural Information Processing Systems 28}.

\bibitem[\protect\citeauthoryear{Rezatofighi \bgroup et al\mbox.\egroup
  }{2019}]{giou}
Rezatofighi, H.; Tsoi, N.; Gwak, J.; Sadeghian, A.; Reid, I.; and Savarese, S.
\newblock 2019.
\newblock Generalized intersection over union: A metric and a loss for bounding
  box regression.
\newblock In {\em The IEEE Conference on Computer Vision and Pattern
  Recognition (CVPR)}.

\bibitem[\protect\citeauthoryear{Song \bgroup et al\mbox.\egroup }{2018}]{TLL}
Song, T.; Sun, L.; Xie, D.; Sun, H.; and Pu, S.
\newblock 2018.
\newblock Small-scale pedestrian detection based on topological line
  localization and temporal feature aggregation.
\newblock In {\em The European Conference on Computer Vision (ECCV)}.

\bibitem[\protect\citeauthoryear{Tian \bgroup et al\mbox.\egroup }{2019}]{FCOS}
Tian, Z.; Shen, C.; Chen, H.; and He, T.
\newblock 2019.
\newblock {FCOS}: Fully convolutional one-stage object detection.
\newblock In {\em The IEEE International Conference on Computer Vision (ICCV)}.

\bibitem[\protect\citeauthoryear{Wang \bgroup et al\mbox.\egroup
  }{2018}]{Wang2018CVPR}
Wang, H.; Wang, Q.; Gao, M.; Li, P.; and Zuo, W.
\newblock 2018.
\newblock Multi-scale location-aware kernel representation for object
  detection.
\newblock In {\em The IEEE Conference on Computer Vision and Pattern
  Recognition (CVPR)}.

\bibitem[\protect\citeauthoryear{Wang \bgroup et al\mbox.\egroup
  }{2019}]{wang2019data}
Wang, H.; Wang, Q.; Yang, F.; Zhang, W.; and Zuo, W.
\newblock 2019.
\newblock Data augmentation for object detection via progressive and selective
  instance-switching.
\newblock {\em arXiv:1906.00358}.

\bibitem[\protect\citeauthoryear{Yang \bgroup et al\mbox.\egroup
  }{2018}]{MetaAnchor}
Yang, T.; Zhang, X.; Li, Z.; Zhang, W.; and Sun, J.
\newblock 2018.
\newblock Metaanchor: Learning to detect objects with customized anchors.
\newblock In {\em Advances in Neural Information Processing Systems}.

\bibitem[\protect\citeauthoryear{Yang \bgroup et al\mbox.\egroup
  }{2019}]{RepPoints}
Yang, Z.; Liu, S.; Hu, H.; Wang, L.; and Lin, S.
\newblock 2019.
\newblock Reppoints: Point set representation for object detection.
\newblock In {\em The IEEE International Conference on Computer Vision (ICCV)}.

\bibitem[\protect\citeauthoryear{Yu \bgroup et al\mbox.\egroup
  }{2016}]{unitbox}
Yu, J.; Jiang, Y.; Wang, Z.; Cao, Z.; and Huang, T.
\newblock 2016.
\newblock Unitbox: An advanced object detection network.
\newblock In {\em Proceedings of the ACM International Conference on
  Multimedia}.

\bibitem[\protect\citeauthoryear{Zhu, He, and Savvides}{2019}]{FSAF}
Zhu, C.; He, Y.; and Savvides, M.
\newblock 2019.
\newblock Feature selective anchor-free module for single-shot object
  detection.
\newblock In {\em The IEEE Conference on Computer Vision and Pattern
  Recognition (CVPR)}.

\end{thebibliography}
	}
	
\end{document}